%% file: main.tex
\documentclass[runningheads]{llncs}

\usepackage{eccv}

\usepackage{eccvabbrv}

\usepackage{graphicx}
\usepackage{booktabs}
\usepackage{epsfig}
\usepackage{amsmath}
\usepackage{amssymb}
\usepackage{lipsum}
\usepackage{bm}
\usepackage{multirow}
\usepackage{tikz}
\usepackage{colortbl}

\usepackage[accsupp]{axessibility}  % Improves PDF readability for those with disabilities.

\usepackage[pagebackref,breaklinks,colorlinks,citecolor=eccvblue]{hyperref}
\usepackage{orcidlink}

\newcommand{\PSNR}{\tiny PSNR$\uparrow$}
\newcommand{\SSIM}{\tiny SSIM$\uparrow$}
\newcommand{\LPIPS}{\tiny LPIPS$\downarrow$}
\makeatletter
\AddToHook{env/tabular/begin}{\let\input\@@input}
\makeatother

\begin{document}

% ---------------------------------------------------------------
% \title{HS-NeRF:\\
\title{Hyperspectral Neural Radiance Fields
\thanks{\scriptsize This work was supported by the National Science Foundation (Award No. 2008302 and Award No. ECCS-2025462); U.S. Department of Agriculture (Award No. 2018-68011-28371); National Science Foundation (Award No. 2112533 and Award No. 1936928); and National Science Foundation-U.S. Department of Agriculture (Award No. 2020-67021-31526)}
}% with Continuous Radiance and Transparency Spectra}


\author{Gerry Chen\inst{1}\orcidlink{0000-0002-1026-1760} \and
Sunil Kumar Narayanan\inst{1} \and
Thomas Gautier Ottou\inst{1} \and
Benjamin Missaoui\inst{1}\orcidlink{0009-0008-9522-8214} \and
Harsh Muriki\inst{1}\orcidlink{0009-0002-1200-8758} \and
Cédric Pradalier\inst{2}\orcidlink{0000-0002-1746-2733} \and
Yongsheng Chen\inst{1}}

\authorrunning{G.~Chen et al.}

\institute{
  Georgia Institute of Technology, Atlanta, GA, USA \and
  Georgia Tech Europe, Metz, France\\
\email{\{gerry,cedric.pradalier\}@gatech.edu}\\
\email{yongsheng.chen@ce.gatech.edu}\\
\url{https://gchenfc.github.io/hs-nerf-website/}}

\maketitle

%%%%%%%%% ABSTRACT
\begin{abstract}
  \input{0-abstract}
\end{abstract}

%%%%%%%%% BODY TEXT
\section{Introduction}

\input{1-intro}

\section{Related Works}

\input{2-related}

\section{HS-NeRF}

\input{3b-approach-hypernerf}

\section{Dataset and Preprocessing}

\input{3a-approach-dataset}

\section{Experiments and Discussion}

\input{4-results_discussion}

\section{Conclusions and Future Works}

\input{5-conclusions_future}

\bibliographystyle{splncs04}
% \bibliography{main}
\bibliography{references/nerf, references/random, references/hyperspectral}
\end{document}

%%%%%%%%% TITLE
% \title{Supplementary Materials for\\~\\HS-NeRF: Hyperspectral Neural Radiance Fields with Continuous Radiance and Transparency Spectra}
\title{Supplementary Materials for \\ Hyperspectral Neural Radiance Fields
\thanks{\scriptsize This work was supported by the National Science Foundation (Award No. 2008302 and Award No. ECCS-2025462); U.S. Department of Agriculture (Award No. 2018-68011-28371); National Science Foundation (Award No. 2112533 and Award No. 1936928); and National Science Foundation-U.S. Department of Agriculture (Award No. 2020-67021-31526)}
}% with Continuous Radiance and Transparency Spectra}


\author{Gerry Chen\inst{1}\orcidlink{0000-0002-1026-1760} \and
Sunil Kumar Narayanan\inst{1} \and
Thomas Gautier Ottou\inst{1} \and
Benjamin Missaoui\inst{1}\orcidlink{0009-0008-9522-8214} \and
Harsh Muriki\inst{1}\orcidlink{0009-0002-1200-8758} \and
Cédric Pradalier\inst{2}\orcidlink{0000-0002-1746-2733} \and
Yongsheng Chen\inst{1}}

\authorrunning{G.~Chen et al.}

\institute{
  Georgia Institute of Technology, Atlanta, GA, USA \and
  Georgia Tech Europe, Metz, France\\
\email{\{gerry,cedric.pradalier\}@gatech.edu}\\
\email{yongsheng.chen@ce.gatech.edu}
% \url{https://hyperspectral-nerf.github.io/supplemental-results-webpage}
}

\maketitle

%%%%%%%%% BODY TEXT
% \tableofcontents

\ifreview
\subsection*{supplemental.html with Qualitative Results}
Please also refer to the \href{./supplemental.html}{supplemental.html} file for qualitative results.  Most operating systems should support double-clicking the html file directly (or ``Open With...'' > ``Google Chrome'') without launching any server processes.  In the event you face issues, you may also access the html 
%at \url{https://hyperspectral-nerf.github.io/supplemental-results-webpage}, whose last updated date can be verified to have been before the ECCV deadline at the anonymized github repo: \href{https://github.com/Hyperspectral-NeRF/supplemental-results-webpage}{Hyperspectral-NeRF/supplemental-results-webpage}.
at \url{https://hyperspectral-nerf.github.io/supplemental-results-webpage}, whose github commit hash 
\href{https://github.com/Hyperspectral-NeRF/supplemental-results-webpage/tree/445006f4fc05d5747e74685a26a8704a7237c6be}{445006f...}
can be verified to have been before the ECCV deadline. 
Some video/image files are also hosted at
\href{https://github.com/Hyperspectral-NeRF/supplemental-results-data}{Hyperspectral-NeRF/supplemental-results-data}
with commit hash
\href{https://github.com/Hyperspectral-NeRF/supplemental-results-data/tree/6205d24cd613f0cfe31ba537ef96352a7c37c3d8}{6205d24...} before the deadline.
\else
\subsection*{Qualitative Results Website}
Please also refer to \url{https://hyperspectral-nerf.github.io/supplemental-results-webpage} for qualitative results. 
\fi

\section{Introduction}
In this work, we demonstrated that Neural Radiance Fields (NeRFs) can be naturally extended to hyperspectral data and are a well-suited tool for hyperspectral 3D reconstruction.  The implementation details provided in this supplemental document describe our simple approach to hyperspectral NeRF, but we anticipate future works by the community will improve upon our baseline implementation using our to-be-published dataset, future larger datasets, additional architecture and hyperparameter tuning, and recent advances in NeRFs.

Our full code will be made publicly available for the camera ready version.

\section{Implementation Details} \label{sec:implementation}
% We openly admit that significant improvements could be made on our implementation, re-emphasizing that our primary contribution is having demonstrated that NeRFs with continuous wavelength representations can work well on hyperspectral data.

\begin{figure}[t]
  \centering
  \includegraphics[width=.9\linewidth]{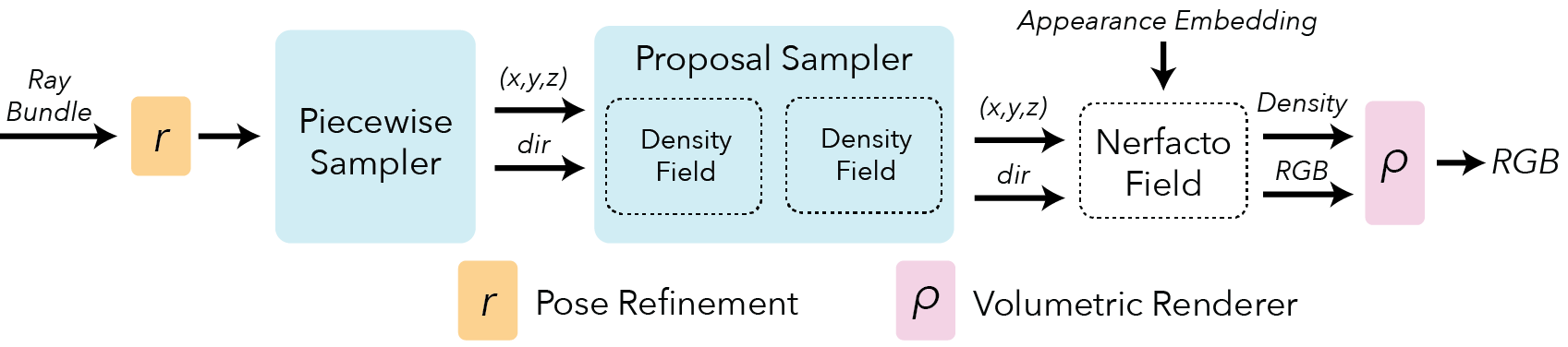}
  \caption{The original nerfacto pipeline (from \href{https://docs.nerf.studio/en/latest/nerfology/methods/nerfacto.html}{nerfstudio docs}) contains a proposal sampler, which is analagous to the ``coarse'' field from the original NeRF paper \cite{mildenhall20eccv_original_nerf}, and a ``Nerfacto Field'', which is analagous to the primary network from the original NeRF paper ($F_\Theta$).}
  \label{fig:nerfacto_pipeline}
\end{figure}

\begin{figure}[t]
  \centering
  \includegraphics[width=.7\linewidth]{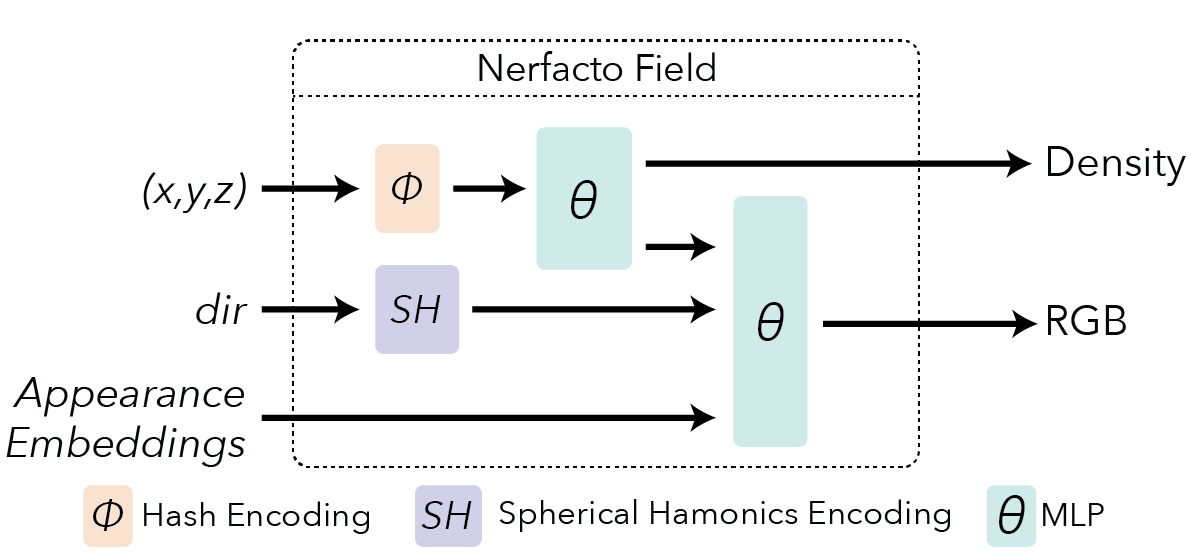}
  \caption{The original nerfacto field (from \href{https://docs.nerf.studio/en/latest/nerfology/methods/nerfacto.html}{nerfstudio docs}) is very similar to the original NeRF paper \cite{mildenhall20eccv_original_nerf}, but includes appearance embeddings \cite{martinbrualla20cvpr_nerfw} and uses slightly different encodings for the position and direction.  This figure is reproduced in Fig. 2 of our main paper.}
  \label{fig:nerfacto_field}
\end{figure}

We build upon nerfstudio's nerfacto implementation, from commit \href{https://github.com/nerfstudio-project/nerfstudio/commit/ef9e00e6141d2fde83fc344b3e27ac475a8597cd}{ef9e00e}.  Our code will be made publicly available for the camera ready paper.  The original nerfacto pipeline and field are shown in Figs. \ref{fig:nerfacto_pipeline} and \ref{fig:nerfacto_field} respectively.


As briefly summarized in the main paper, we make relatively minimal modifications to the pipeline and field.  Using the notation from Section 5.4: Ablations, $\bm{C}_1$ only changes the rightmost MLP in Fig. \ref{fig:nerfacto_field} to output 128 channels in the last layer instead of 3; $\bm{C}_2$ changes the positional hash encoding ($\phi$ in Fig. \ref{fig:nerfacto_field}) to take 4 inputs instead of 3 (appending $\lambda$) and changes the rightmost MLP to only have 1 output for $c$ instead of $(r,g,b)$; and $\bm{C}$ is shown in Fig. 2 (bottom) of the main paper.  For $\bm{C}$, the sinusoidal encoding for $\lambda$ is taken to have 8 terms (tested 2, 4, 8, 16 terms, with 8 performing marginally better than 4 and 16, and 2 significantly worse).  Also for $\bm{C}$, the component $\bm{C}(\lambda; \Theta_C)$ MLP from Fig. 2 of the main paper was taken to be identical to the rightmost MLP in Fig. \ref{fig:nerfacto_field} except with the appropriate additional number of inputs to accommodate concatenating the sinusoidally encoded wavelength, and with only 1 output for $c$ instead of 3 for $(r,g,b)$.  The latent vector $\Theta_C$ was taken to be the same size as in the nerfacto implementation (15-dim), with increasing the size to 32 and 64 showing negligible performance improvement but increased training instability.

Similarly, $\sigma_0$ is the stock nerfacto field (scalar); $\sigma_1$ only changes the left MLP in Fig. \ref{fig:nerfacto_field} to have 128 outputs; $\sigma_2$ changes the positional hash encoding to take 4 inputs, and $\sigma$ is as shown in Fig. 2 (bottom) of the main paper.  The additional component $\sigma(\lambda;\Theta_C)$ MLP has 3 layers with 64-dim hidden layers and ReLU activations.  The sinusoidally encoded $\lambda$ is shared with $\bm{C}$ and the latent $\Theta_\sigma$ vector is shared with (identical to) the $\Theta_C$ vector.

Finally, $P_0$ is the stock nerfacto proposal network while $P_\lambda$ augments the proposal network with the wavelength.  For $P_\lambda$, the position is first run through a hash encoding and MLP as in $P_0$, except the MLP outputs a latent vector of dimension 7 instead of a scalar density.  This latent vector is concatenated with a 2-term sinusoidally encoded wavelength and fed through a 2-layer network with 7-dim hidden layer to output a scalar density for inverse transform ray sampling.  Like the original nerfacto pipeline, this sampling step occurs twice with identical architecture (but different weights) proposal networks.

Reiterating our implementation, our primary HS-NeRF implementation uses $\bm{C}(\lambda; \Theta_C)$, $\sigma_0(\lambda; \Theta_\sigma)$, and $P_0$, which we find to produce good results while also enabling wavelength interpolation.

\subsection{RGB Implementations} \label{ssec:rgb}
% \paragraph{Erratum.} First, we apologize for the following error in the main paper that we will correct for the camera-ready version:
% For the caption in Table 1, we mistakenly state that our method outperforms the baseline for the more challenging Tools and Origami scenes.  We intended to portray that our approaches achieve very comparable performance to standard RGB nerfacto on all scenes, despite the fact that, for Ours-Cont and Ours-RGB, the wavelength bands are very far apart which should make learning \emph{more} difficult.  For Ours-Hyper, we achieve comparable performance on all scenes except Tools despite the fact that we are learning 128 channels instead of just 3 while the number of learnable parameters is virtually identical to nerfacto and completely identical to Ours-Cont.

\paragraph{Pseudo-RGB wavelengths.} For the purposes of generating pseudo-RGB images, on the Surface Optics datasets we use the wavelengths 622nm, 555nm, and 503nm for R, G, and B channels respectively.

For the BaySpec datasets, we use a slightly more involved approach.  We found that the BaySpec datasets were more sensitive to noise saturation and white balance, so we use an approach similar to that described in Section 6.2 of the main paper to generate pseudo-RGB images.  Specifically, we first manually identify 5-10 point correspondences between a hyperspectral image and an iPhone photo of the same scene to represent pairs of colors that should be the same.
Expressing the $n$ points in the hyperspectral image as $X \in \mathbb{R}^{128\times n}$ and in the iPhone photo as $Y \in \mathbb{R}^{3\times n}$, we solve for a linear transformation $A \in \mathbb{R}^{3\times 128} = \arg\min_{A'} \norm{Y-A'X}^2$ using the least squares solution.  We then use this transformation to convert the hyperspectral image to pseudo-RGB.  After using this initial approach to boot-strap certain components of the pipeline, we later apply the method described in Section 6.2 to generate pseudo-RGB renderings.

% Generating more accurate pseudo-RGB images by integrating over the spectrum according to an image sensor sensitivity curve (as described in Section 5.5 of the main paper) would also be possible, but is unnecessary to demonstrate our results.

\paragraph{HS-NeRF RGB variation implementations.} For the purposes of making a quantitative comparison to standard RGB NeRF, Section 5.2 and Table 1 of the main paper present variations of our approach applied to just 3-channel (RGB) images instead of the full 128-channel hyperspectral data.  As described in the caption of Table 1, ``Ours-Cont'' refers to our HS-NeRF implementation but trained on only 3 wavelengths (so we maintain a continuous representation for radiance spectra, but have very weak supervision of only 3 channels), ``Ours-RGB'' refers to $\bm{C}_1, \sigma_1, P_0$ with 3 output channels for both $\bm{C}_1$ and $\sigma_1$, and ``Ours-Hyper'' refers to our HS-NeRF implementation trained on all 128 wavelengths.  In the table for Ours-Hyper, PSNR and SSIM are evaluated over all 128 wavelengths while LPIPS is evaluated on the RGB images obtained using the Pseudo-RGB procedure.

\section{Training Details} \label{sec:training}
All networks were trained for 25000 steps, with 4096 train rays per step using the Adam optimizer.
The proposal networks and field both used lr=1e-2, eps=1e-15, and an exponential decay lr schedule to 1e-4 after 20000 steps.  Camera extrinsic and intrinsic optimization were both turned off, since evaluation metrics are skewed if camera parameters are modified.  To accommodate imperfect camera poses, after COLMAP, stock nerfacto was run on Pseudo-RGB images for 100000 steps with camera optimization turned on and the resulting camera pose corrections were saved and used in subsequent tests.  The Surface Optics datasets took roughly 20 minutes to train HS-NeRF while the BaySpec datasets took roughly 40 minutes to train on an RTX 3090 due to the need to re-cache a new set of 32 images every 50 steps (see next paragraph).  Most architectures required similar training times, with the exception of the last two rows of the ablation: $\bm{C}_2 \sigma_2 P_0$ and $\bm{C} \sigma P_\lambda$ took at roughly three times as long.

For the Surface Optics datasets, of the 48 images per image set, 43 were used for training and 5 withheld for evaluation.  Each step, the 4096 training rays were sampled randomly from all 43 training images, except for row 6 of the ablations where the training rays were sampled from only 10 of the 43 training images each step, with the choice of 10 images being re-sampled every 50 steps.
The BaySpec datasets were too large to fit in VRAM so rays were sampled from 32 images every step, with the set of 32 images being re-sampled every 50 steps, with row 6 of the ablations being reduced to 12 images resampled every 50 steps.

In some approaches, not all wavelengths could be run for every step due to VRAM limits so a subset of wavelengths were sampled (randomly) for each step, but every sampled wavelength was run for every ray in the step.  For rows 1 and 2 of the ablations, every wavelength could be run every step.  For rows 3, 4 (HS-NeRF, ours), and 5, the number of wavelengths sampled per step were 8, 12, and 6, respectively.

For evaluation, every wavelength of every pixel of the 5 (Surface Optics) or 35 (BaySpec) evaluation images were evaluated and compared for each scene.

\begin{figure}
  \centering
  \includegraphics[width=.9\linewidth]{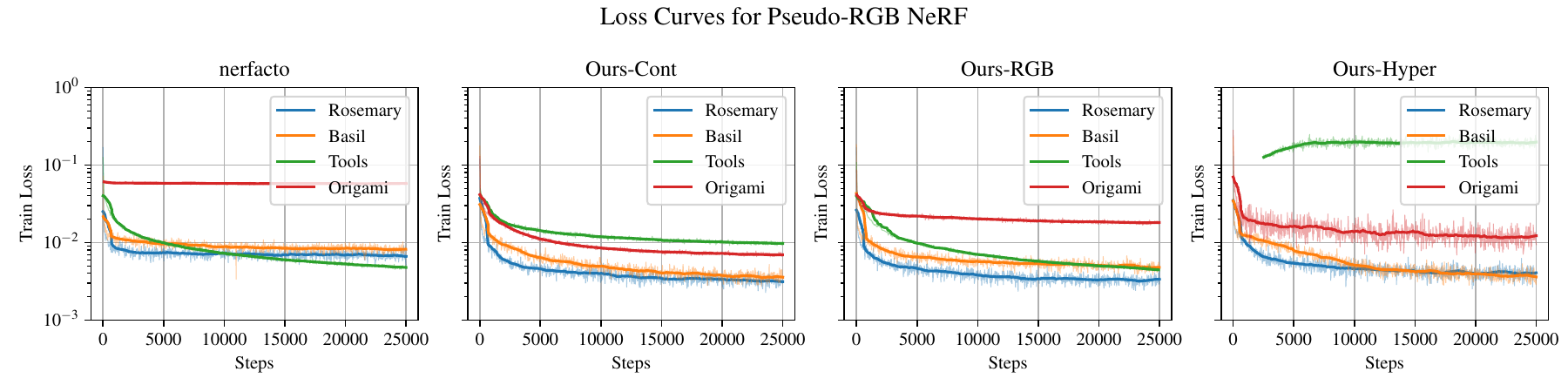}
  \caption{Loss curves for RGB NeRF correspond to the metrics from Table 1 in the main paper.  Most scenes have converged by 25000 steps except the Tools scene which appears to have difficulty converging for all methods except ``Ours-Cont''}%, which is reflected in Table 1 of the main paper.}
  \label{fig:loss_rgb}
% \end{figure}
% \begin{figure}
  \centering
  \includegraphics[width=0.8\linewidth]{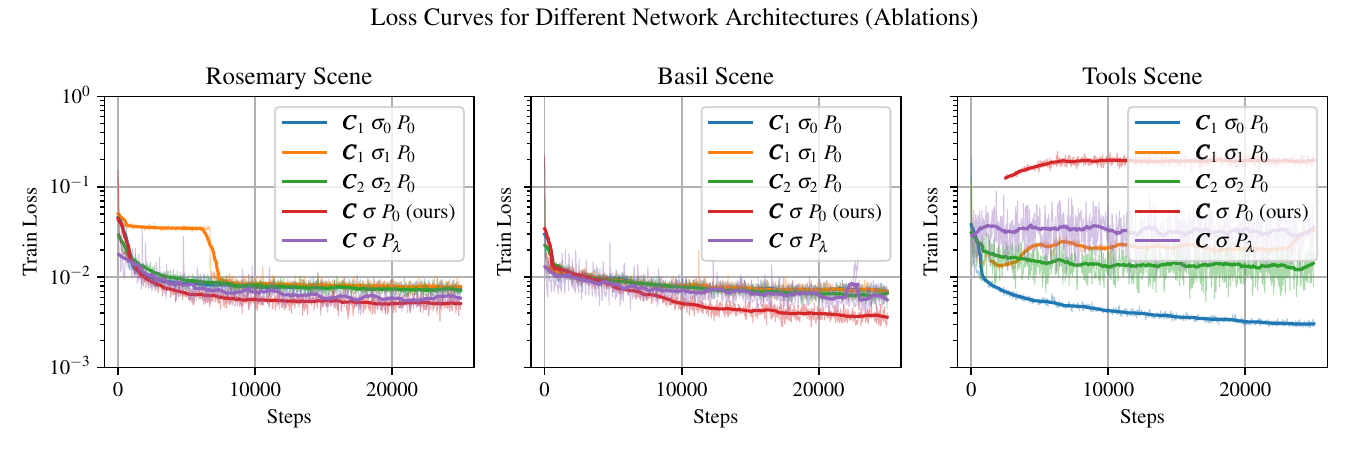}
  \caption{Loss curves for ablation testing (analagous to Table 3 in the main paper) shows that while the rosemary and basil scenes optimize well, the tools scene does not converge particularly well for any method, re-emphasizing the suspected pre-processing (COLMAP) inaccuracy.\\}
  \label{fig:loss_ablations}
% \end{figure}
% \begin{figure}
  \centering
  \includegraphics[width=.4\linewidth]{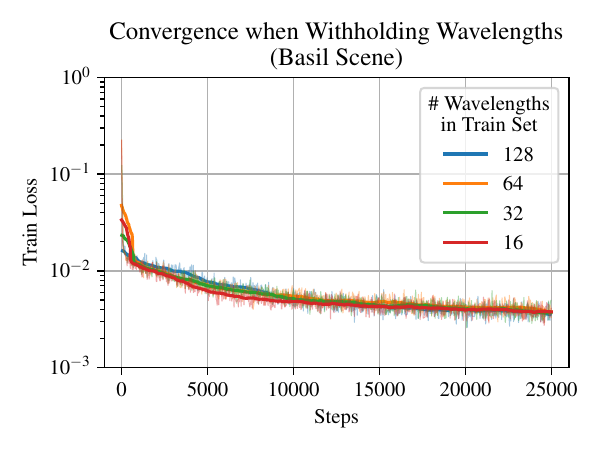}
  \caption{Loss curves for HS-NeRF trained with a subset of wavelengths (analogous to Table 2 in the main paper) shows that even training with only 1 out of every 8 wavelengths still has almost identical convergence rate w.r.t. number of steps.}
  \label{fig:loss_numwavelengths}
\end{figure}

\subsection{Commentary on the Tools Scene}
The Tools scene experienced instabilities during training with several approaches including both HS-NeRF (ours) and nerfacto (RGB baseline).
We anticipate that obtaining better camera intrinsics and extrinsics may correct this issue, since (a) every method had difficulty on this scene and (b) enabling camera pose optimization during NeRF training improved convergence for all methods.
Better camera intrinsics could be obtained by initializing COLMAP with the intrinsics obtained from other scenes, and better camera extrinsics could be obtained through a combination of tuning COLMAP parameters, utilizing turntable priors, and a longer NeRF-based camera pose refinement as described in \ref{sec:training}.
The poor convergence on the Tools scene for all methods is illustrated in both Fig. \ref{fig:loss_rgb} (green curves) and Fig. \ref{fig:loss_ablations}.

\subsection{Loss Curves}
To demonstrate that all methods were fairly trained until convergence, the loss curves corresponding to some metrics given in the main paper are shown.  As mentioned, the Tools scene appears to have difficulty converging for all methods including baseline nerfacto, suggesting possible pre-processing (COLMAP) inaccuracy.  This is evident both in the green curves of Fig. \ref{fig:loss_rgb} and in the rightmost plot of Fig. \ref{fig:loss_ablations}.  Evidencing the hyperspectral super-resolution (spectral interpolation) application, Fig. \ref{fig:loss_numwavelengths} shows almost identical training loss for all subsets of wavelengths trained with.

% Exemplified by Fig. \ref{fig:loss_ablations} (left, orange), one interesting observation we found is that the $\bm{C}_1$ and $\sigma_1$ architectures occasionally exhibit convergence followed by a second descent and convergence.  Inspecting the evaluation images, we observe the first convergence to be learning a scalar density field and the second convergence to be learning the color spectrum.

\section{Qualitative Example Results}
\ifreview
A selection of example images and videos with brief explanations are provided on the \href{./supplemental.html}{supplemental.html} page in the enclosing zip folder to better gauge our results qualitatively.

Most operating systems should support double-clicking the html file directly (or ``Open With...'' > ``Google Chrome'') without launching any server processes.  In the event you face issues, you may also access the html at \url{https://hyperspectral-nerf.github.io/supplemental-results-webpage}, whose github commit hash 
\href{https://github.com/Hyperspectral-NeRF/supplemental-results-webpage/tree/445006f4fc05d5747e74685a26a8704a7237c6be}{445006f...}
can be verified to have been before the ECCV deadline. 
Some video/image files are also hosted at
\href{https://github.com/Hyperspectral-NeRF/supplemental-results-data}{Hyperspectral-NeRF/supplemental-results-data}
with commit hash
\href{https://github.com/Hyperspectral-NeRF/supplemental-results-data/tree/6205d24cd613f0cfe31ba537ef96352a7c37c3d8}{6205d24...} before the deadline.
\else
A selection of example images and videos with brief explanations are provided at \url{https://hyperspectral-nerf.github.io/supplemental-results-webpage} to better gauge our results qualitatively.
\fi

\ifreview
\clearpage
\fi
{\small
\bibliographystyle{ieee_fullname}
\bibliography{references/nerf, references/random, references/hyperspectral}
}

%% file: 0-abstract.tex
Hyperspectral Imagery (HSI) has been used in many applications to non-destructively determine the material and/or chemical compositions of samples.
% While 2D images can give point or aggregate measurements, a
% However, the sensitivity of HSI to lighting and viewing conditions poses challenges for wider scale applications.
There is growing interest in creating 3D hyperspectral reconstructions, which could provide both spatial and spectral information while also mitigating common HSI challenges such as non-Lambertian surfaces and translucent objects.
However, traditional 3D reconstruction with HSI is difficult due to technological limitations of hyperspectral cameras.
In recent years, Neural Radiance Fields (NeRFs) have seen widespread success in creating high quality volumetric 3D representations of scenes captured by a variety of camera models.
% it is typically only used to create a single aggregate measurement of the subjects as opposed to a spatially varying description.
%
% In this work, we extend Neural Radiance Fields (NeRF) from RGB to hyperspectral data to compute hyperspectral 3D reconstructions of scenes from images taken with a hyperspectral camera and turntable. 
%
% Meanwhile, NeRFs have recently seen widespread success creating high quality 3D representations of scenes from images.
Leveraging recent advances in NeRFs, we propose computing a hyperspectral 3D reconstruction in which every point in space and view direction is characterized by wavelength-dependent radiance and transmittance spectra.
%
% We present a novel NeRF-based approach to predict continuous emittance and transmittance spectra instead of scalar volume density and 3-dimensional color intensity.
To evaluate our approach, a dataset containing nearly 2000 hyperspectral images across 8 scenes and 2 cameras was collected.  We perform comparisons against traditional RGB NeRF baselines and apply ablation testing with alternative spectra representations. 
Finally, we demonstrate the potential of hyperspectral NeRFs for hyperspectral super-resolution and imaging sensor simulation.
We show that our hyperspectral NeRF approach enables creating fast, accurate volumetric 3D hyperspectral scenes and enables several new applications and areas for future study. 
% Hyperspectral imagery is useful as a non-destructive data collection technique in many applications because measuring reflectance, transmittance, and/or fluorescence at different wavelengths can indicate the material or even the concentrations of different molecules.
% 3D reconstruction from images is also useful as a non-destructive data collection technique because inferring the 3D structure of an object can reveal 
% \patrick{Honestly, really nice sounding abstract. Nailed the academic tone. I'm not sure if I'd say "we present three approaches", since that's confusing. I'd present one approach and some ablations.}

%% file: 1-intro.tex
% \patrick{Remember that your audience is a bunch of deep learning script kiddies who never touch hardware. I didn't know what hyperspectral was until two years ago, and I'd bet most of them don't know either. You might have a figure of some sort to clearly state what hyperspectral is. I can imagine either a photo of a plant or a diagram of the EM spectrum}

Hyperspectral imagery is a useful tool in many applications for non-destructively characterizing material and chemical compositions.
For example, HSI is used in agriculture to assess plant health and nutrient content, in medicine to diagnose diseases, and in drilling to view otherwise invisible gasses like methane.
In contrast to typical RGB images which have 3 color channels for each pixel, hyperspectral images consist of tens to hundreds of color channels (wavelengths) for each pixel and typically have minimal spectral overlap among channels.
Because different materials and molecules have different reflectance, transmittance, and/or fluorescence properties at different wavelengths, hyperspectral data may be used to infer the composition of a sample.
% blah blah hyperspectral.
However, studying the \textit{spatial} data in HSI is currently under-studied for a number of reasons, with many works only using the ``image'' part of HSI to select individual point or aggregated foreground pixel statistics.
% \patrick{You might list some application domains of hyperspectral. Plants, modelling emission gases, humans, idk}

\begin{figure}
    \centering
    \includegraphics[width=0.49\linewidth,page=4,trim={0 2.13in 4.13in 2.935in},clip]{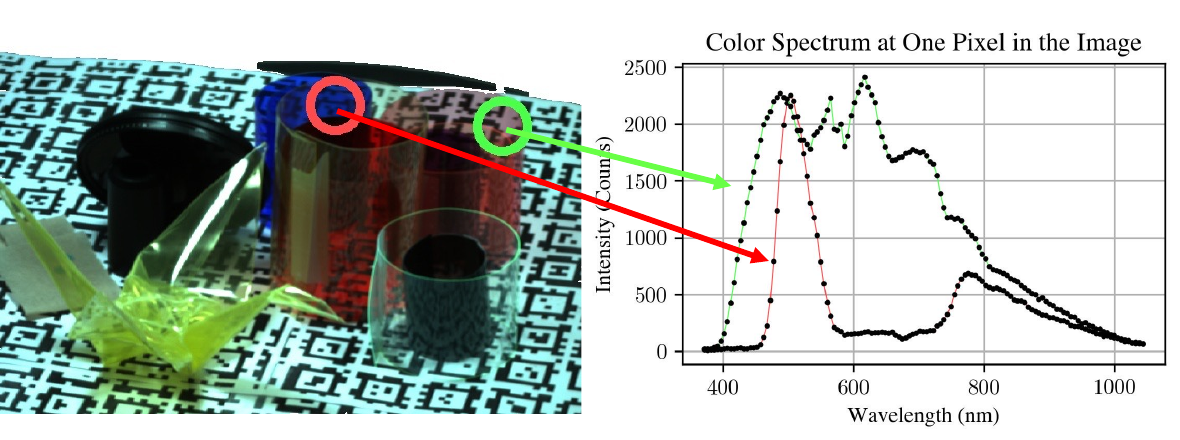}
    \includegraphics[width=0.49\linewidth,page=4,trim={0 5.065in 4.13in 0},clip]{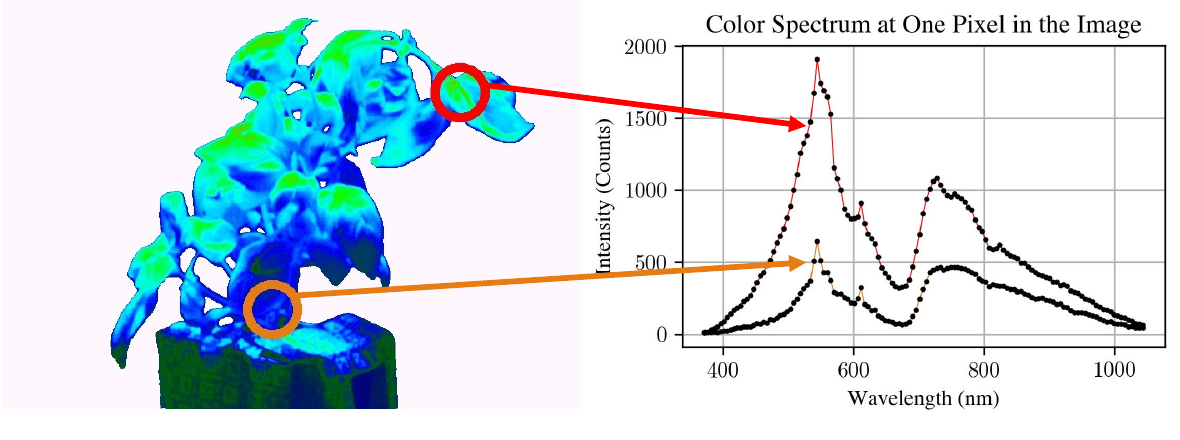}
    \caption{Instead of 3 color channels for each pixel, hyperspectral images have many color channels per pixel to measure the color spectrum for every pixel.  In this work, we leverage recent advances in Neural Radiance Fields to create hyperspectral 3D scene representations.  Left: an example hyperspectral image and spectrums at 2 points.  Right: a hyperspectral NeRF and spectra.}
\end{figure}

We believe NeRF-based 3D reconstructions address many challenges unique to hyperspectral data.
% We believe creating NeRF-based 3D reconstructions of hyperspectral data may help alleviate many issues associated with leveraging spatial hyperspectral information.
Illumination angle dependence and low signal-to-noise ratio (SNR) can both be mitigated by \emph{fusing} information from many images from different viewpoints.  The volumetric radiance field representation also provides a continuous spatial interpolation, in contrast to the sparse point-cloud representations in traditional SfM or multi-view stereo approaches.
In our approach, we also show how we can use wavelength as a \emph{continuous input} to the NeRF allowing interpolation of not only position and view angle, but also of wavelength to enable hyperspectral super-resolution.  Finally, we believe NeRF-based approaches may be able to handle non-Lambertian surfaces, partial transparency, and wavelength-dependent transparency better than SfM approaches.

Our contributions are as follows:
\begin{itemize}
    \item Collect and share a \textbf{dataset} of hyperspectral images suitable for hyperspectral 3D reconstruction,
    \item Identify \textbf{special considerations} needed to accommodate hyperspectral camera limitations when computing NeRFs,
    \item Introduce our \textbf{HS-NeRF model} for HyperSpectral 3D reconstruction,
    % with {continuous representations for radiance and transmittance spectra}
    with evaluations and ablations,
    \item \textbf{Demonstrate the feasibility} of creating hyperspectral 3D reconstructions using NeRFs, and
    \item Demonstrate potential \textbf{applications} of hyperspectral NeRFs.
\end{itemize}

%% file: 2-related.tex
\subsubsection*{Hyperspectral Imagery.} \label{ssec:related_hs}
We defer to the many high-quality review papers on detailed background and applications of HSI such as \cite{hyperspectral_review}.  However, we briefly motivate the need for hyperspectral 3D reconstruction and discuss some practical considerations of hyperspectral cameras.

Challenges that have been identified in hyperspectral literature include the black-box nature of correlating spectra with sample properties \cite{hyperspectral_review}, the low signal to noise ratio \cite{Liang13iccv_plant_hyperspectral_3dmodeling}, and the high cost and inconvenience of high-resolution hyperspectral imaging.  We believe that fusing multiple hyperspectral images into a 3D model can help scientists develop more mechanistic understandings of hyperspectral data and improve the signal to noise ratio.  Further, we believe many recent advances surrounding NeRFs, such as NeRF in the Dark \cite{mildenhall2022cvpr_raw-nerf} and Deblur-NeRF \cite{Ma2021arxiv_Deblur-NeRF}, may help extract more information with cheaper HSI sensors.

3D reconstruction is particularly difficult due to a few undesirable properties of HSI cameras.
% Although we will not discuss the technical details, there are a few undesirable properties which are common to nearly all hyperspectral cameras.
First, there is a tradeoff between spatial, spectral, and temporal (exposure time) resolution such that obtaining a low-noise, high-resolution image with many wavelength bands will necessarily require a long (typically on the order of minutes) exposure time.
Second, lenses for hyperspectral cameras are limited in power due to the wavelength-dependent index of refraction of glass (even IR-corrected glass is not perfect), which creates more exaggerated chromatic aberration and increases the cost of optics.  Many hyperspectral cameras have extremely narrow fields of view as a result, while still suffering from inconsistent focus across wavelengths and narrow depths of field.
% hyperspectral cameras \colorbox{red}{have narrow fields of view and} are incompatible with standard lenses due to the wavelength-dependent index of refraction of glass.
Finally, aperture size is typically bounded due to interactions with order-blocking filters which correct diffraction side-effects, further limiting the ability to capture clear images in varying environments.
We discuss how we address these challenges for our dataset and approach.
% First, there is a tradeoff between spatial resolution, spectral resolution, and exposure time (temporal resolution).  For a fixed image sensor size and aperture, increasing the resolution of one property typically requires sacrificing another.  Another common trait is a narrow field of view which is caused by a combination of mechanical limitations and glass lenses.  The index of refraction of glass is wavelength dependent which prohibits strong lensing.  Even IR-corrected glass in not perfect.  Finally, 

% Ranging from handheld single-point hyperspectral sensors \cite{gong2002analysis} to satellite-based hyperspectral analysis of entire fields \cite{}, hyperspectral imaging has been widely applied for non-destructive analysis in many industries.  For example, in agriculture, hyperspectral imagery can be used to estimate plant and soil nutrient content because, much like different materials have different visual appearances in RGB, different molecules have different appearances in IR wavelengths.  Ranging from handheld single-point hyperspectral sensors \cite{gong2002analysis} to satellite-based hyperspectral analysis of entire fields \cite{}, hyperspectral imaging 

% \begin{itemize}
%   \item hyperspectral camera mechanics: space-time-wavelength resolution tradeoff, FOV, chromatic aberration
%   \item applications of hyperspectral imagery / prior works
% \end{itemize}

\subsubsection*{Hyperspectral 3D Reconstruction.}
Creating hyperspectral 3D reconstructions from hyperspectral images has been attempted in the past with point cloud-based methods.  \cite{Zia15wacv_hs_3d_reconstruction} creates separate point clouds for each wavelength channel then merges them to create a hyperspectral point cloud while \cite{Liang13iccv_plant_hyperspectral_3dmodeling} directly performs Structure-from-Motion on the hyperspectral data.  Extending this, \cite{Ma22access_hyperspectral_keypoints_descriptors} designs custom hyperspectral keypoint feature descriptors for hyperspectral images to aid in 3D reconstruction, while several other works also address hyperspectral features for image classification \cite{Kumar20tf_feature_extraction}.  However, Structure-from-Motion approaches often generate only sparse point clouds and hyperspectral imagery may often be too noisy and low resolution to obtain good multi-view stereo results.  Further, point clouds typically do not provide sufficient occupancy information to accurately compensate for shadows and lighting variations.
\cite{Xu20osa_hyperspectral_structured_light_projector} takes a different approach and designs a hyperspectral structured light project device to measure 3D hyperspectral information.  Somewhat similarly, \cite{Kim12acm_3d_hyperspectral_project} projects hyperspectral images onto existing 3D geometry models.  However, these are not as flexible or scalable as a camera-only solution.

\subsubsection*{Neural Radiance Fields.}

Neural Radiance Fields (NeRFs) have \emph{exploded} \cite{Dellaert20blog_nerf_explosion} in popularity since the original paper by Mildenhall et al. was published \cite{mildenhall2022cvpr_raw-nerf}.  NeRFs present a deep-learning approach to obtaining a high quality 3D representation of a scene by learning a function mapping the location of a point in space and the direction from which it is being viewed to color radiance and volume density.  To determine the color a pixel of an image should take, a rendering step queries the function along the pixel's corresponding image ray and composites the colors according to classical volume rendering \cite{mildenhall2022cvpr_raw-nerf}.  A large body of works has since extended and improved upon the initial NeRF paper.

Although no works to our knowledge directly tackle the hyperspectral 3D reconstruction problem using NeRFs, we directly leverage several advancements such as the substantial efficiency improvements from Instant-NGP \cite{mueller2022instant} and the open-source nerfstudio package and nerfacto implementation \cite{nerfstudio} which we build our implementation upon.  We also draw inspiration from many related works.  For example, several spatio-temporal \cite{Xian20cvpr_space-time-nerf,Du20arxiv_nerflow_4d-time-nerf}, deformable, and other NeRF works \cite{Gao21cvpr_novelviewsynthesis_time} append a scalar time variable to the 3D location input  similar to an approach we compare against concatenating wavelength to location.  Similarly, Zhi et al.'s semantic NeRF work using implicit scene representations for semantic super-resolution \cite{Zhi21iccv_semantic_nerf_implicit_scene_representation} inspires our continuous wavelength representation for hyperspectral super-resolution.

Several works could also complement our work well and we hope future research can incorporate their techniques for HS-NeRF.  For example, RawNeRF \cite{mildenhall2022cvpr_raw-nerf} and NAN \cite{pearl2022noiseaware-NAN} both leverage NeRF's information fusing ability for low-light denoising which could help reduce the exposure time required.  RawNeRF applies post-processing on the NeRF instead of the input photos, which could be applied to mitigate artifacts of hyperspectral cameras such as order-blocking filter interference.  AR-NeRF \cite{Kaneko2022cvpr_AR-NeRF} and Deblur-NeRF \cite{Ma2021arxiv_Deblur-NeRF}, which address depth of field/defocus and motion blur, respectively, could also be useful given the long exposure times and aperture limitations of hyperspectral cameras.

\subsubsection*{Hyperspectral Super-Resolution.}
Evidenced by numerous papers, datasets \cite{Akhtar15cvpr_hs_superresolution_fusion}, and competitions \cite{Arad18cvprw_ntire_hyperspectral_superresolution_challenge}, the hyperspectral super-resolution task has become increasingly popular.  Hyperspectral super-resolution may refer to obtaining more wavelength resolution (\ie use an RGB or multi-spectral image to predict a hyperspectral image), obtaining more spatial resolution (\ie use a low-resolution hyperspectral image to predict a higher resolution one), or more commonly fusing together information from complementary sensors \cite{Han22acm_hyperspectral_multiscale_fusion,Akhtar15cvpr_hs_superresolution_fusion}.  Perhaps the most similar to this work is \cite{Zhang21arxiv_implicit_neural_hs_superresolution} which uses an implicit neural representation to predict a higher resolution image using a continuous function mapping pixel coordinate to color.  We extend their work to 3D and put it in the context of NeRFs.

We are also proud to publish our dataset of almost 2000 hyperspectral images; one plausible reason for the relatively greater popularity of hyperspectral super-resolution over hyperspectral 3D reconstruction is the lack of publicly available datasets for the latter.

In summary, we believe our work is highly complementary to existing works and supports a promising new direction in 3D hyperspectral reasoning research.

%% file: 3b-approach-hypernerf.tex
% \patrick{IMO CV people like naming their approaches. In the title you picked HS-NeRF, which is cool. You might plug that in here as many times as possible, so this section header is HS-NeRF, and many of the refs in the text can be replaced with HS-NeRF}
% We build our implementation of HS-NeRF upon nerfstudio's ``nerfacto'' implementation \cite{nerfstudio}.

\begin{figure}
  \centering
  \includegraphics[width=0.49\linewidth,page=1,trim={0 5.93in 5.8in 0},clip]{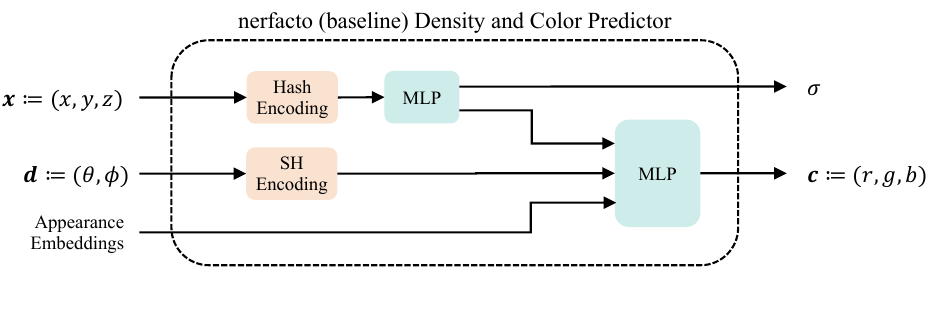}\hfill
  \includegraphics[width=0.49\linewidth,page=1,trim={0 3.86in 5.8in 2.07in},clip]{figs/slides.pdf}
  \caption{
    To handle hyperspectral data, we include a wavelength input to our network which predicts a scalar color intensity and a scalar transmittance.  The network produces spectra for the color intensity and transmittance via the latent vectors $\bm{\Theta}_C$ and $\bm{\Theta}_\sigma$, respectively, and networks $\bm{C}(\lambda;\bm{\Theta}_C)$ and $\sigma(\lambda;\bm{\Theta}_\sigma)$ compute the value of the spectra at the queried wavelength.
    % In designing HS-NeRF, we discuss and compare 3 different approaches to modifying ``nerfacto'' \cite{nerfstudio} for hyperspectral prediction involving both the color radiance and transmittance prediction networks.  We find that Approach 1 is best on most scenes, but Approaches 2 and 3 outperform when image quality is low.
    % We base our implementation of HS-NeRF on ``nerfacto'' \cite{nerfstudio}.
    % The sampling and rendering components remain largely unchanged.
    % Since we do not change other aspects of the architecture (\eg sampler \& render per wavelength), shown above is just the differences between the networks that predict radiance and density given position, view direction, and optionally appearance embeddings and wavelength.  The left plot is the baseline nerfacto design while the bottom is our HS-NeRF design.
    }
  \label{fig:fields}
\end{figure}

Building on the ``nerfacto'' \cite{nerfstudio} NeRF implementation, we discuss 3 modifications for HS-NeRF to accomodate hyperspectral data: (1) color radiance prediction, (2) transmittance spectrum prediction, and (3) proposal network modification.
% We build HS-NeRF upon ``nerfacto'' \cite{nerfstudio}.
% Figures \ref{fig:overall_network} and \ref{fig:fields} illustrate the overall architecture (same as nerfacto) and the modified subcomponent of the network, respectively.

As compared to a $(H, W, 3)$ RGB image, a hyperspectral image can be represented by a $(H, W, N)$ tensor, where $H$ and $W$ are the height and width of the image, and $N$ is the number of channels/wavelengths.

Instead of directly predicting an $N$-dimensional color radiance, we choose to represent color radiance and transmittance both as continuous spectra: functions of wavelength.  We do so by predicting latent vectors which represent parameters of learned spectral plots which can then be evaluated for a given wavelength as shown in Fig. \ref{fig:fields}.

% In our final implementation, we represent both color emission and color absorption as continuous spectra -- intensities as functions of wavelength -- by predicting latent vectors representing parameters of learned spectral plots which are evaluated for a given wavelength with decoders.
% In our final implementation, we represent both color emission and color absorption as continuous spectra -- intensities as functions of wavelength -- by predicting latent vectors $\bm{\Theta}_c, \bm{\Theta}_\sigma$ representing parameterizations of learned spectral plots which are evaluated with decoders $d_c(\lambda~;~\bm{\Theta}_c) = c^\lambda$ and $d_\sigma(\lambda~;~\bm{\Theta}_\sigma) = \sigma^\lambda$.

In this section, we describe the math formulations and some implementation details, though we also discuss and compare alternatives in Section \ref{ssec:ablations}.  For additional details, please refer to the supplemental materials.
% , as well as several alternative architectures tested in our ablations, for predicting color emission and absorption spectra, as well as a ray sample proposal network that predicts a wavelength-dependent proposal density.
% We introduce 3 primary changes to the nerfacto architecture: (1) the color emission network is changed to output a color spectrum, (2) the color absorption network is changed to output a wavelength-dependent absorption spectrum, and (3) the proposal network is changed to output a wavelength-dependent proposal spectrum.\\

% \colorbox{red}{I think Fig2 is too small/too hardh to read}

\subsection{Color Radiance Spectrum Prediction}
% Whereas the nerfacto network predict a 3-channel color vector, we must be able to predict a color radiance spectrum.  For HS-NeRF, we use $\bm{c}_1$: the color component of Approach 1, in which we simply expand the last layer of the color radiance network to output a 128-dimensional vector, where each element represents the color intensity at a different wavelength.  We choose 128 wavelengths to match the number of channels in our hyperspectral images, but this number can be adjusted to match the number of channels in the hyperspectral images being used.  In other words, we parameterize the radiance spectrum by a 128-dimensional sampling according to the camera's sensor bands.
% choose to predict a continuous color radiance spectrum.
% We also discuss several alternative approaches and compare their performance in the ablations section (\ref{ssec:ablations}).

% In addition to this simple approach, we also discuss and test two other approaches which 

We predict the continuous radiance spectrum by first predicting a latent vector $\bm{\Theta}_C$ representing the parameters of a learned spectral plot.  We then obtain the radiance $c^\lambda$ for a given wavelength $\lambda$ by passing the latent vector together with the a sinusoidal positionally encoded wavelength $\lambda$ through a decoder $\bm{C}$. %(\lambda~;~\bm{\Theta}_C) = c^\lambda$.
Formally, whereas the nerfacto (baseline) network outputs the color intensity on a ray as:
\begin{align}
  \bm{C}_0 : (\bm{x}, \bm{d}) \rightarrow \bm{c} := (r,g,b)
\end{align}
where $\bm{x}:=(x,y,z)$ and $\bm{d}:=(\theta,\phi)$ are the location and view direction of the ray, respectively, we predict the color radiance spectrum as:
\begin{align}
  \bm{C} : (\lambda ; \Theta_C(\bm{x}, \bm{d})) \rightarrow c^\lambda
\end{align}
where
  % $c^\lambda$ denotes the scalar emitted (color) intensity at wavelength $\lambda$,
  % $\lambda$ denotes the query wavelength,
  % $\Theta_c(\bm{x}, \bm{d}): \mathbb{R}^5 \rightarrow \mathbb{R}^{n_\Theta}$ is a network that maps the ray's location and view direction to a latent vector $\bm{\Theta}$, and
  $\Theta_c(\bm{x}, \bm{d})$ is a network that maps the ray's location and view direction to a latent vector $\bm{\Theta}_c$.
  %  with dimensionality $n_\Theta$.

% Implementation details are provided in \ref{ssec:implementation}.

\subsection{Color Transmittance Spectrum Prediction}
Similarly, the transmittance spectrum describes a wavelength-dependent volume density.
In other words, instead of using a scalar density field to describe the transparency of the scene, we investigate the possibility of using a wavelength-dependent density field.

Although wavelength-dependent transmittance can also be applied to RGB scenes and isn't strictly necessary for hyperspectral scenes, it is generally more interesting for hyperspectral imagery due to the fact that many materials are transparent in visible wavelengths but opaque in IR or vice-versa.

In the original and nerfacto NeRF implementations, the volume density is given by a scalar function $\sigma(\bm{x})$.
Instead, we choose to model the volume density in much the same way as for color radiance: a network $\Theta_\sigma(\bm{x})$ predicts a latent vector $\bm{\Theta}_\sigma$ which is passed with the wavelength to another network
\begin{align}
  \sigma : (\lambda;\Theta_\sigma(\bm{x})) \rightarrow \sigma^\lambda  
\end{align}
where $\sigma^\lambda$ denotes the density at wavelength $\lambda$.

\subsection{Wavelength-dependent Proposal Network} \label{ssec:wavelength-dependent-proposal-network}
Finally, when choosing a wavelength-dependent volume density, it may also be natural to make the sample proposal network (analagous to the ``coarse'' network) wavelength-dependent.
Including such a dependence may be especially useful in larger scenes with many partially transparent objects (such as plastic films).
However, in ablations with our dataset, we found that doing so did not improve performance and caused more training instability.  %Nevertheless, we encourage others to try since it is likely dataset dependent.
Furthermore, as we/nerfacto use the proposal loss from Mip-NERF 360 \cite{barron22cvpr_mipnerf360} which encourages the coarse loss to be an upper bound of the fine-level density network, the proposal network is penalized when it under-estimates \textit{any} wavelength's density.

%% file: 3a-approach-dataset.tex
Before being able to train NeRF models on hyperspectral images,
we first collect images using a hyperspectral camera and turntable, apply preprocessing, and obtain camera poses and intrinsics by running COLMAP on pseudo-RGB images.

% There are 3 primary challenges we face in this approach to applying NeRF techniques to hyperspectral imagery: (1) collecting and preprocessing images given hyperspectral camera limitations, (2) applying NeRF to non-ideal images, and (3) extending RGB NeRF techniques to handle additional wavelengths.

% To address the first challenge, we first create ``pseudo-RGB'' images from the hyperspectral images by simulating an RGB image sensor's spectral sensitivity and determine the modifications necessary to generate RGB NeRF models.
% To address the second challenge, we test 3 different approaches.

\subsection{Data Collection Setups}
\begin{figure}[h!]
  \centering
  \begin{minipage}[t]{.45\textwidth}
    \centering
    \includegraphics[width=1.0\linewidth]{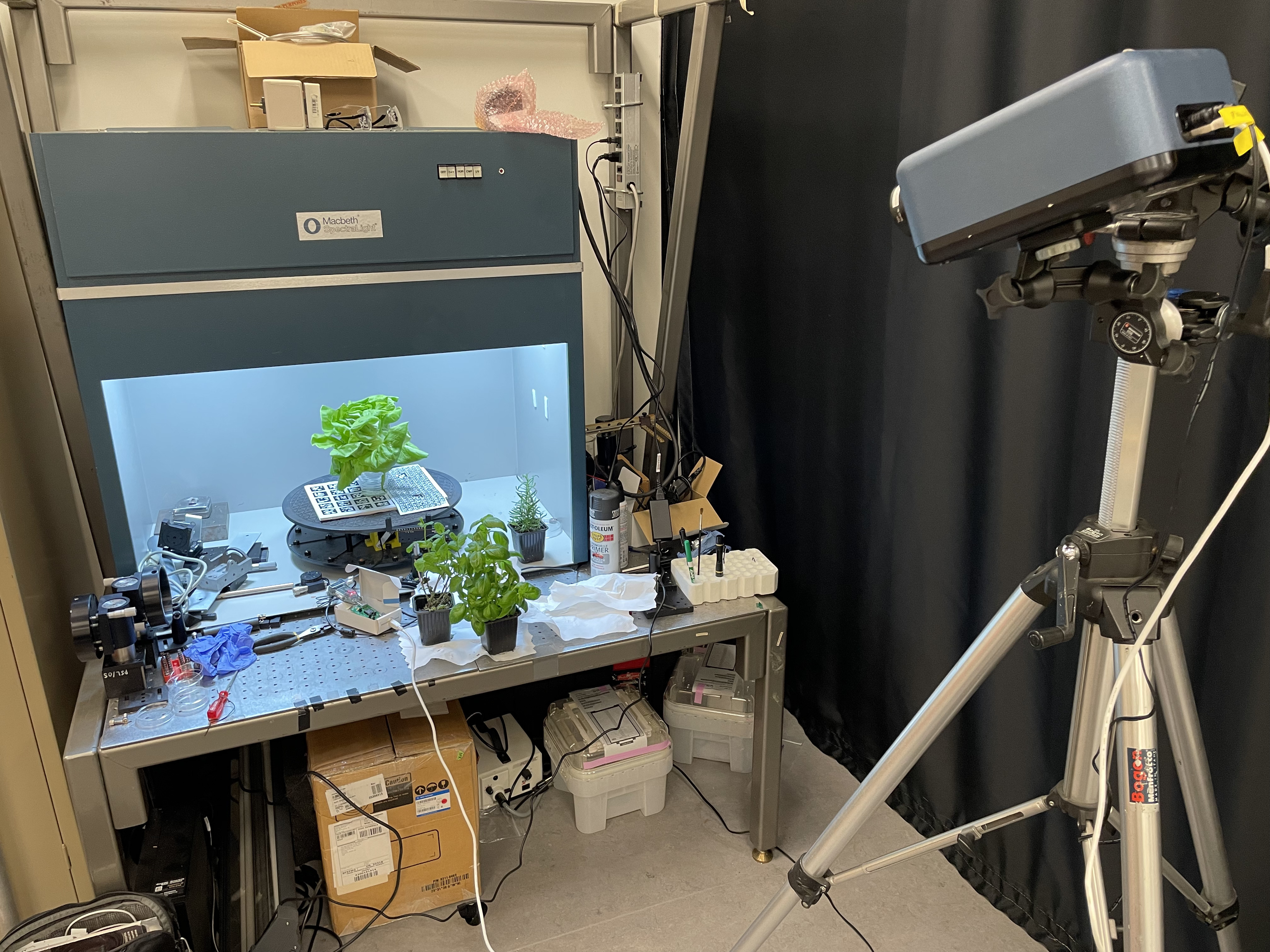}
    \caption{
      % The Surface Optics SOC710-VP camera imaging setup.
      The Surface Optics SOC710-VP camera is mounted on a tripod and the sample of interest is placed on a turnable in a Macbeth SpectraLight lightbooth with a light gray background.  The camera is roughly 2 meters away from the scene due to its shallow depth of field and narrow field of view.
    }
    \label{fig:exp_setup}
  \end{minipage}\hfill%
  \begin{minipage}[t]{.45\textwidth}
    \centering
    \includegraphics[width=0.78\linewidth]{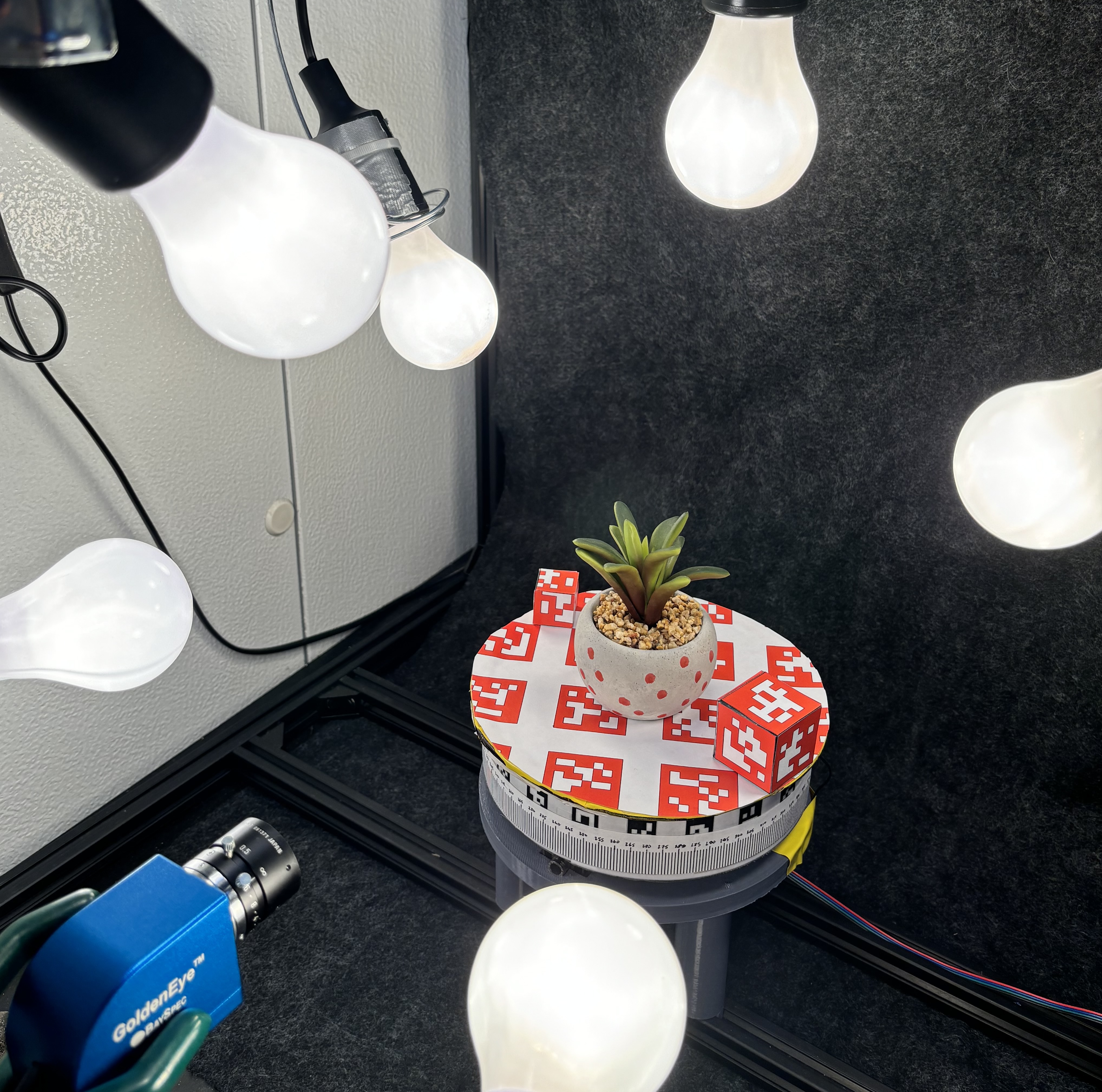}
    \caption{
      % The BaySpec GolderEye camera setup.
      The BaySpec GoldenEye camera is held with a laboratory clamp and the sample of interest (here, an \textit{Anacampseros} plastic plant) is placed on a turnable under tungsten halogen lighting. The camera is roughly 20 centimeters away from the scene thanks to its wide field of view.
    }
    \label{fig:exp_setup2}
  \end{minipage}%
\end{figure}

To demonstrate the generalizeability of our approach, we compose our dataset of scenes using two different hyperspectral cameras.  The imaging setups are shown in Figs \ref{fig:exp_setup} and \ref{fig:exp_setup2}, respectively, and full details on the cameras and data collection procedure are available in the supplemental materials.

\begin{figure}
  \centering
  \includegraphics[width=0.8\linewidth]{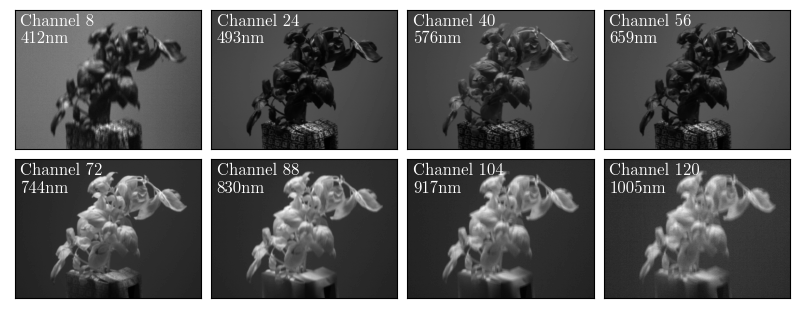}
  \caption{By visually inspecting the same image of the basil plant in several different wavelengths, it becomes obvious that the additional information afforded by HSI makes background removal significantly easier than in RGB images.}
  \label{fig:hs_background_removal}
\end{figure}

\subsection{Image Acquisition and Preprocessing}
As mentioned in Section~\ref{ssec:related_hs}, hyperspectral cameras have inherent non-idealities that must be accounted for when collecting data. % that primarily revolve around (1) the tradeoff between spatial, temporal, and spectral resolution, and (2) the wavelength-dependent refractive index of lens materials.
We focus our discussion on the two following hyperspectral cameras:

(a) The Surface Optics SOC710-VP has a high spatial ($696\times520$ pixels) and spectral resolution ($N=128$) at the expense of poor temporal resolution (long exposure time) and extends from 370nm to 1100nm causing some wavelength-dependent refractive index effects (blurry and misaligned in far-IR).

(b) The BaySpec GoldenEye camera also has a high spatial ($640\times512$ pixels) and spectral  ($N=140$) resolution over the range 400 -- 1100nm. Unlike the other camera, this one is of the "snapshot" type, meaning that an hypercube can be captured \textit{almost} instantly, in around 3 seconds, but at the expense of significant grain and noise.

First, for both cameras, interactions with glass lenses and diffraction gratings necessitate careful choice of aperture size and lens.  
In short, the wavelength-dependent refractive index of glass (even for IR-corrected lenses) necessitates a small aperture to keep all wavelengths in focus while diffraction effects necessitate a large aperture to satisfy the criteria for the order-blocking filter commonly used in hyperspectral cameras.
In response, the Surface Optics camera uses a pre-calibrated 35mm lens with F5.6 aperture, and we place it around 2 meters from the scene to both increase the depth of field and accommodate the narrow field of view of the lens.
We find that far-IR wavelengths are slightly out of focus (\eg Fig \ref{fig:hs_background_removal}) and, although they are not particularly problematic in this work, techniques from \cite{Kaneko2022cvpr_AR-NeRF,Ma2021arxiv_Deblur-NeRF} may be used.
The BaySpec camera, on the other hand, has a pre-calibrated 8mm lens of  40° field of view that we use with a F16 aperture and place at just 20 centimeters from the scene.

% Wavelength-dependent refractive index effects cause an effect analogous to chromatic aberration in RGB cameras, whereby not all wavelengths can be simultaneously in focus and misalignment among wavelengths may occur due to the glass lens focusing different wavelengths differently even for IR-corrected lenses.  Reduced aperture size can mitigate this effect, but increases diffraction effects and interferes with the order-blocking filter commonly used in hyperspectral cameras 

% Second, due to chromatic aberration effects, lenses are very limited so the field of view of the camera is very narrow.  Talk about canonicalizing the camera poses and cropping the scene / ray sampler to the FOV.  Also talk about ability of NeRF to ``cheat'' due to small overlap.

Second, the image backgrounds do not rotate with the scene so they must be removed from the images.
Fortunately, background removal is straightforward when leveraging hyperspectral data, as illustrated in Figure \ref{fig:hs_background_removal} where water in the paint is highly reflective in IR compared to the painted background.
We set the background color to pure white: 255 (Surface Optics) or pure black: 0 (BaySpec) in all wavelength channels. %\patrick{maybe a bit long on the details, but I'm sure it'll get cleaned up later}

\subsection{Computing Camera Poses and Scene Bounds} \label{ssec:colmap}
To compute the camera intrinsics and extrinsics necessary to train NeRF models, we create grayscale images to use in COLMAP: an off-the-shelf Structure-from-motion package.
% Although the turntable enforces the angular position of the camera in a ``ring'' around the scene, the distance, height, and orientation of the camera are unknown.
% We use COLMAP, a popular Structure-from-Motion package, to obtain camera poses and intrinsics.
%
% We generate pseudo-RGB images to feed into COLMAP by taking the 3 wavelength channels from the hyperspectral images which roughly correspond to red, green, and blue.  Although a more accurate pseudo-RGB image could be generated,
% %as in Figure \ref{fig:simulated_cameras}
% we find the narrower wavelength bands create more distinctive features and thus more reliable matching.
We generate the gray-scale images by selecting the channel with the greatest foreground intensity variance.
Due in part to the narrow field of view (Surface Optics), low resolution, chromatic aberration (Surface Optics), and grainy noise (BaySpec) compared to \eg smartphone cameras, we need to use an undistorted pinhole camera model (distortion parameters caused poor optimization results), have many high-quality features in the scene (which we achieve using AprilTags \cite{krogius2019iros_AprilTag}), and apply a strict matching threshold (inlier ratio $\ge$ 0.70, \# inliers $\ge$ 25).

Finally, as a byproduct of the narrow field of view of the Surface Optics camera, we also find it imperative to crop the ray sampler tightly to the scene to avoid sampling points that are only visible in a few cameras.  Failing to do so results in ``cheating'' whereby the NeRF model synthesizes many 2D ``screens'' in front of each camera outside the field of view of the other cameras instead of a single consistent 3D object.
% To determine suitable ray sampling bounds, we canonicalize the camera poses according to Figure \ref{fig:canonicalization} and compute the ``scene'' bounding box, which describes the ray sampler's bounds, by projecting the cameras' fields of view onto the $xz$ and $yz$ planes.
To determine suitable ray sampling bounds, we canonicalize the camera poses and compute the ``scene'' bounding box, which describes the ray sampler's bounds, by projecting the cameras' fields of view onto the $xz$ and $yz$ planes (see Figure in Supplemental).
%into a common coordinate frame centered at the origin with the lowest ``ring'' of camera poses forming a circle centered on the z-axis with radius 1 and placed at an elevation such that the cameras are oriented pointing at the origin.  We then estimate the scene size by projecting the camera's field of view onto the $xz$ plane (for a camera on the y-axis).

% \begin{figure}
%   \centering
%   \includegraphics[width=0.5\linewidth,page=2,trim={2.75in 0.65in 2.75in 0.75in},clip]{figs/slides.pdf}
%   \caption{To tightly bound the scene to the objects of interest, we canonicalize the camera poses as shown and compute a bounding box centered at the origin whose size is determined by the camera's field of view.}
%   \label{fig:canonicalization}
% \end{figure}

% Due to the aforementioned limitations of hyperspectral cameras, we find 3 primary artifacts that are not particularly commonplace in NeRF applications: (1) the background is not rotating with the scene, (2) the scene is not centered in the image, and (3) the scene is not in focus.

\subsection{Dataset Scenes}

We collect the following datasets, depending on the camera:

(a) With the Surface Optics camera, we collect a dataset of 4 scenes of 48 images each, with 2 of the scenes exhibiting intricate plant geometry (``Rosemary'' and ``Basil'') and the other two exhibiting several objects with wavelength-dependent transparency and radiance/reflection (``Tools'' and ``Origami'').  Given the hyperspectral camera's strength in measuring wavelength and comparative weakness at capturing spatial resolution, we expect the latter two scenes to be more challenging.

(b) With the BaySpec camera, we collect a dataset of 4 scenes of 433 images each, all exhibiting intricate plant geometry  (``\textit{Anacampseros}'' and ``\textit{Caladium}'' made of plastic, a ``Pinecone'', and a "Buttercrunch Lettuce").

%% file: 4-results_discussion.tex
We train HS-NeRFs on the 8 scenes from our dataset and compare the results both quantitatively and qualitatively (see supplemental).  We evaluate the reconstruction accuracy of the network compared to a nerfacto baseline and run an ablation.  We also present sample applications of hyperspectral super-resolution and camera sensor simulation.

\subsection{Evaluation Metrics}

\subsubsection*{Validation Set.} The validation set is formed the standard way as in the NeRF literature: 90\% of the images in the dataset are used to train and the other 10\% used for validation by comparing the actual image with the NeRF prediction at the given camera pose.
\subsubsection*{Metrics.}
As is standard in NeRF literature, we present the Peak Signal to Noise Ratio (PSNR), the Structural Similarity Index Measure (SSIM), and the Learned Perceptual Image Patch Similarity (LPIPS) metrics.
Note that, for LPIPS, we use pseudo-RGB images extracted as described in \ref{ssec:simulating_imaging_sensors}: as the integral of the product between RGB spectral sensitivity curves and the pixel's intensity spectrum.
% For ablations, we also compare training speeds.
% \todo{note: extract MSE from loss dict}
In addition to quantitative metrics, we also provide a qualitative comparison of synthesized images.

\subsection{RGB} \label{ssec:RGB}
We can first evaluate our hyperspectral approach on RGB images using stock nerfacto as a baseline.  This is possible since a standard RGB image can be interpreted as an  $N=3$-channel hyperspectral image, and our approach can generalize to any number of channels.  Instead of using standard RGB datasets, we test using pseudo-RGB images generated from our dataset as described in \ref{ssec:simulating_imaging_sensors} to maintain the noise, aberration, and other effects present in the original hyperspectral images.

% However, because the wavelengths corresponding to $(r, g, b)$ are very sparse as compared to hyperspectral images, the assumption that the color and density spectra are continuous is violated.  Therefore, in addition to our approach using $F_{2b}$, we also present results using $F_1$ for both the emission and absorption networks.
In addition to training our HS-NeRF network directly with 3 wavelengths (``Ours-Cont'') and training a stock nerfacto model (``nerfacto''), we also consider two other comparisons.  First, we train a modified version of the network in \textit{Row 2} of the ablations to output 3 discrete radiance and 3 discrete density channels (``Ours-RGB'').  Second, we provide the metrics for a HS-NeRF network trained and evaluated on the full, 128-channel hyperspectral image (``Ours-HS''), which serves to show that extending from RGB to hyperspectral does not significantly degrade performance.  For Ours-HS, We evaluate LPIPS on pseudo-RGB images extracted as described in \ref{ssec:simulating_imaging_sensors}.

% ``Ours-Cont'' refers to the exact same implementation as HS-NeRF except we only train with 3 wavelengths.  ``Ours-RGB'' refers to a slightly modified version of row 2 in the ablations (see Table \ref{tab:ablations}) to output 3 discrete radiance and 3 discrete density channels.  ``Ours-HS'' refers to our HS-NeRF implementation trained on a full, 128-channel hyperspectral image.  Since our metrics are normalized to the number of wavelengths, we provide it as a reference to evidence that the hyperspectral performance on any given channel is comparable to the performance for RGB channels.

Our hyperspectral approach outperforms the baseline in almost every category.  This is extremely pronounced for the BaySpec datasets (bottom) which have significant noise in the ground truth images.  As a result, the model trained on the full hyperspectral data is more robust by virtue of the additional information given across different frames and wavelengths - similar to how camera calibration and 3D computer vision can often achieve sub-pixel accuracy.

\begin{table}
  \centering
  \caption{RGB Results.
  Our HS-NeRF approach outperforms the NeRF baseline (nerfacto) on almost every baseline when evaluating on RGB images (N=3 wavelengths).
  }
  % Because our approach can handle arbitrary numbers of wavelengths, we can apply it to the RGB case (N=3 wavelengths) to compare with a baseline NeRF implementation (nerfacto).
  % We observe that our hyperspectral approach outperforms the nerfacto baseline in almost every category.
  \label{tab:rgb}
  % \small
  \scriptsize
  Surface Optics Datasets
  \input{tables/rgb}\\[0.5em]
  BaySpec Datasets\\
  \input{tables/rgb_bayspec}
\end{table} % We are pretty far over the page limit so I'm trying to be a bit stingy with space XD ;)sorry :)

\subsection{Ablations} \label{ssec:ablations}

Instead of the HS-NeRF network described in Fig. \ref{fig:fields}, we also compare against simpler approaches to constructing hyperspectral NeRFs.  We investigate the following simplifications:
\begin{enumerate}
    \item The transmittance function is not wavelength dependent (scalar).
    \item Instead of inputting the wavelength, the network simply outputs 128 channels (instead of 3 for RGB).
    \item The wavelength is instead input as another spatial dimension similar to the way time is handled in time-varying NeRFs \cite{Du20arxiv_nerflow_4d-time-nerf,Xian20cvpr_space-time-nerf}.
\end{enumerate}
We also investigate making the proposal networks wavelength-dependent as discussed in Sec. \ref{ssec:wavelength-dependent-proposal-network}.

% As mentioned in previous sections, there are some alternative options for how to achieve hyperspectral radiance and density predictions.  We may:
% \begin{enumerate}
%   \item simply output $N$-dimensional vectors instead of a scalar density or 3-channel color,
%   \item input the wavelength as another spatial dimension similar to the way time is handled in time-varying NeRFs \cite{Du20arxiv_nerflow_4d-time-nerf,Xian20cvpr_space-time-nerf},
%   \item keep a grayscale density, and/or
%   \item augment the proposal networks (coarse networks) with the wavelength with the same options, or leave the proposal network as is.
% \end{enumerate}

\newcommand{\C}{\bm{C}}
\newcommand{\xd}{\bm{x},\bm{d}}

We denote the options for the radiance spectrum as:
% \begin{align*}
% \C_1(\xd)\rightarrow (c^\lambda_1,\ldots, c^\lambda_N) &&\hspace*{2em}&&
% \C_2(\lambda,\xd)\rightarrow c^\lambda &&\hspace*{2em}&&
% \C(\lambda; \bm{\Theta}_C(\xd)),
% \end{align*}
% where in $\C_2$, $\lambda$ is concatenated with $\bm{x}$ before the hash encoding.
% We denote the options for the the transmittance spectrum as $$
$$\def\arraystretch{1.2}
\begin{array}{cl@{\hspace{0.1em}:\hspace{0.1em}}c@{~\rightarrow~}l}
  \text{(ours)}     & \bm{C}   & (\lambda~;~\bm{\Theta}_c(\bm{x}, \bm{d})) & c^\lambda  \\
  % \text{(nerfacto)} & \bm{C}_0 & (\bm{x}, \bm{d}) & \bm{c} := (r, g, b) \\
                    & \bm{C}_1 & (\bm{x}, \bm{d}) & (c^{\lambda_1}, \ldots, c^{\lambda_N}) \\
                    & \bm{C}_2 & (\lambda, \bm{x}, \bm{d}) & c^\lambda ,
\end{array}
$$
where in $\bm{C}_2$, $\lambda$ is concatenated with $\bm{x}$ before the hash encoding.

Similarly, we denote the options for the density spectrum as:
$$\def\arraystretch{1.2}
\begin{array}{cl@{\hspace{0.1em}:\hspace{0.1em}}c@{~\rightarrow~}l}
  \text{(ours)}     & \sigma   & (\lambda~;~\bm{\Theta}_\sigma(\bm{x})) & \sigma^\lambda  \\
                    & \sigma_0 & (\bm{x}) & \sigma \\ % \in \mathbb{R}\\
                    & \sigma_1 & (\bm{x}) & (\sigma^{\lambda_1}, \ldots, \sigma^{\lambda_N}) \\
                    & \sigma_2 & (\bm{x}, \lambda) & \sigma^\lambda.
\end{array}
$$

Finally, for the proposal network we only consider $P_0$, which denotes baseline nerfacto network, and $P_\lambda$, which denotes a proposal network augmented with the wavelength.  

The ablation results are shown in Table \ref{tab:ablations} and a representative sample shown in Figure \ref{fig:ablation}.
% All methods are comparable, with the exception of the last two rows on the Tools scene.  This is believed to be a byproduct of imperfect COLMAP camera pose computation which resulted in a camera pose offset from the correct pose.  Given that the LPIPS is not negatively affected and the SSIM is affected less than the PSNR, it is reasonable to expect that an image translation could produce such an error, especially given these are real datasets and not synthetic.
We first observe that the continuous representations perform consistently well.  Meanwhile, the discrete approaches excel at the noisy images from the BaySpec camera but struggle with the cleaner (but fewer) images from the Surface Optics scenes.
Although not quantitatively represented, $\C_2,\sigma_2$ is significantly harder to train, very frequently diverging and taking 3 times as long on average due to the fact that full passes through the network must be evaluated for every wavelength.
We also observe that the choice of transmittance function has little effect on the results, with Rows 1-2 matching well and Rows 3-4 matching well.
Qualitatively (see supplemental), we can also observe that all methods except the wavelength-dependent proposal network perform very well.
Finally, we observe that our approach appears to have overall the best performance.

% \todo{move this to implementation section} Implementation-wise, $E_1$ uses the same network as in the nerfacto model except the last layer of the color prediction MLP outputs 128 channels instead of 3.  $E_{2a}$ uses the same network as in the nerfacto model except the wavelength is appended to the direction vector before the spherical encoding.  Finally, $E_{2b}$ uses the same network as in the the nerfacto model except the color prediction network outputs an $n_\Theta$-dimensional latent vector $\bm{\Theta}$ instead of a 3-channel color vector.  The latent vector is then appended to a 4-frequency sinusoidal encoded wavelength and decoded by a 2-layer MLP that outputs the color intensity at a given wavelength.

% The different ways of describing 

% To study the effects of various ways to parameterize 

% Of the $3\cdot 4\cdot 4 = 48$ possible combinations of the color emission ($F_1, F_{2a}, F_{2b}$), color absorption ($F_0, F_1, F_{2a}, F_{2b}$), and wavelength-dependent proposal network ($F_0, F_1, F_{2a}, F_{2b}$) architectures,  focus on only the following 5 combinations:

\begin{table*}
  \centering
  \caption{Ablations.}% : We can see that on low noise scenes like Rosemary, Basil, and Tools, the early integration of wavelength information ($C_2$) yields better results. Whereas on more noisier scenes, like \textit{Anacampseros}, \textit{Caladium}, and Pinecone, predicting the discrete Hyperspectral channels achieves better results. ($C_1$) }
  \scalebox{0.8}{
    \input{tables/ablation}
  }
  \\[0.5em]
  \scalebox{0.8}{
    \input{tables/ablation_bayspec}
  }
  \label{tab:ablations}
\end{table*}

\input{figs/abblation_grid}

% \subsection{SfM baselines?}
% No time to implement, and they don't fit with NeRF metrics anyway.

\section{Example Applications} \label{ssec:applications}
The ability to represent a scene with a radiance field that is continuous not only in position and view direction but also in wavelength opens up a variety of applications which we very briefly demonstrate here.

\subsection{Hyperspectral Super-Resolution}
% HS-NeRF naturally enables hyperspectral super-resolution given its continuous spatial and spectral representations.

Hyperspectral super-resolution, in which we seek to (a) turn a \emph{multi}spectral image (with fewer wavelengths than a hyperspectral image) into a hyperspectral image (with more wavelengths) or (b) turn a low-resolution hyperspectral image into a higher resolution image, is an increasingly popular challenge in computer vision.  Zhang et al. has already applied a similar continuous spectral network representation to 2D hyperspectral super-resolution \cite{Zhang22icme_neural_hs_superresolution}, and leveraging multi-view consistency may further improve the performance of existing hyperspectral super-resolution approaches.

\input{4_results_superresolution}
\subsection{Simulating Imaging Sensors} \label{ssec:simulating_imaging_sensors}
HS-NeRF can also characterize \textit{and} simulate different image sensors from a single photo of the imaged scene.
Since camera image sensors each have a particular spectral response, the recorded RGB intensities are given by integrating (over all wavelengths) the product of the light spectrum reaching the sensor and the sensor's response curve.  HS-NeRF can compute the light spectrum reaching the sensor at each pixel which enables both characterization and simulation.

\begin{figure}
    \centering
    \includegraphics[width=0.5\linewidth]{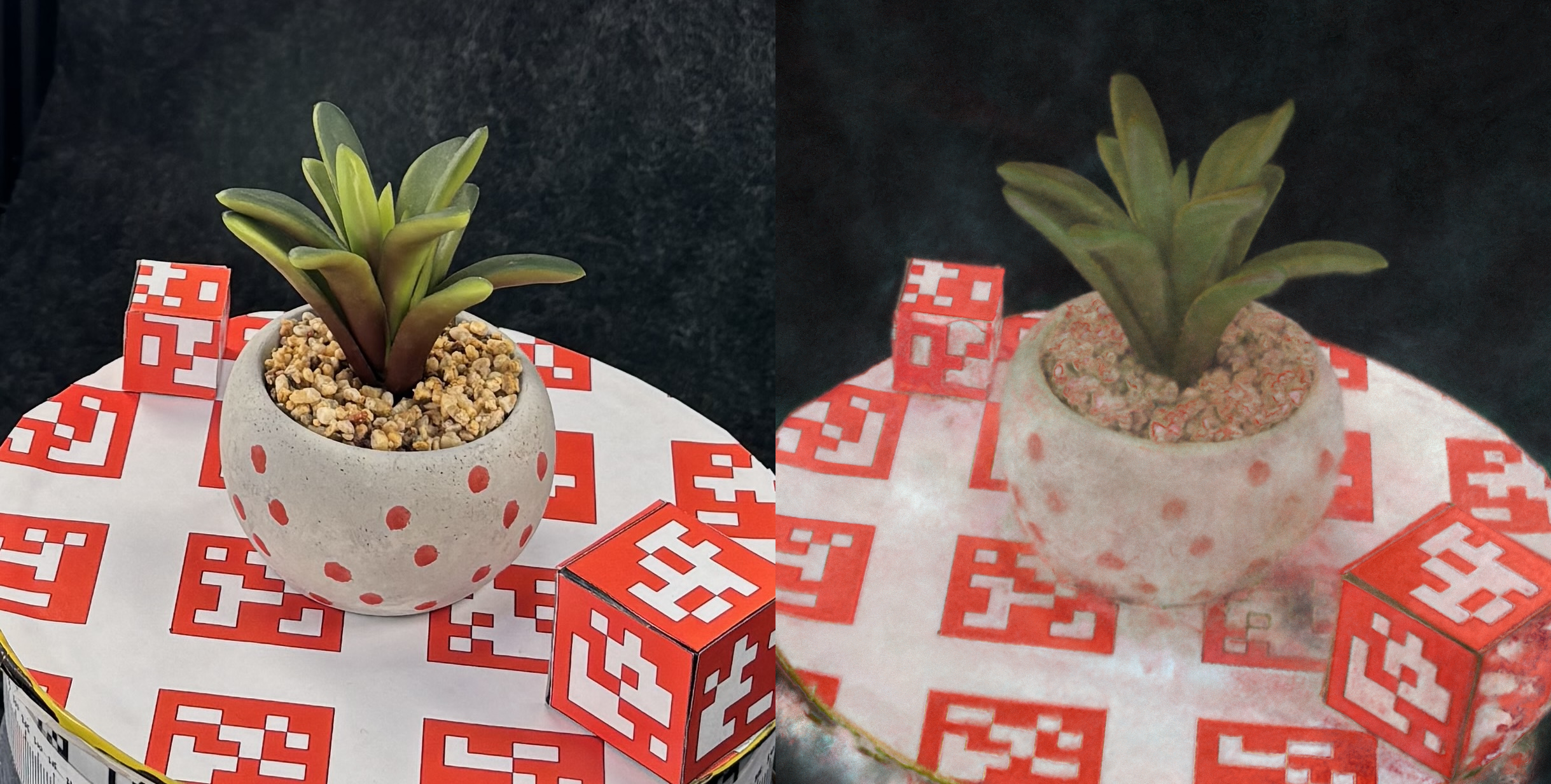}
    \caption{A demonstration of image sensor simulation (here with the \textit{Anacampseros}).  Left: A photo taken with a smartphone (a crop from Fig. \ref{fig:exp_setup2}).  Right: The simulated photo computed from the HS-NeRF.}
    \label{fig:camera_simulation}
\end{figure}

To characterize the image sensor, we first localize the query photo's camera pose in the scene using COLMAP.  We then render a hyperspectral image from the HS-NeRF.  Finally, we solve for the spectral response by minimizing:
\begin{align}
    \left[\bar{r}(\lambda)~~\bar{g}(\lambda)~~\bar{b}(\lambda)\right] &= \arg\min_X \sum_{i,j} (RGB_{ij} - X \cdot HS_{ij})^2
\end{align}
where $X\in \mathbb{R}^{3 \times 128}$; $RGB_{ij} \in \mathbb{R}^3$ and $HS_{ij} \in \mathbb{R}^{128}$ denote the $ij^{th}$ pixel of the photo and hyperspectral render, respectively; and $\bar{r},\bar{g},\bar{b}$ denote 128x1 vector parametrizations of the image sensor's spectral response, assuming each pixel has the same spectral response.
To simulate an image sensor, we simply apply the response: $RGB_{simulated, ij} = \left[\bar{r}(\lambda)~~\bar{g}(\lambda)~~\bar{b}(\lambda)\right] \cdot HS_{ij}$.  Fig. \ref{fig:camera_simulation} shows an example simulated photo alongside the real photo.

%% file: tables/rgb.tex
\begin{tabular}[htp]{l ccc ccc ccc ccc} \toprule
  \multirow{2}{*}{Method} & \multicolumn{3}{c}{Rosemary} & \multicolumn{3}{c}{Basil} & \multicolumn{3}{c}{Tools} & \multicolumn{3}{c}{Origami} \\
           & \PSNR        & \SSIM & \LPIPS & \PSNR           & \SSIM & \LPIPS & \PSNR & \SSIM & \LPIPS & \PSNR & \SSIM & \LPIPS \\
  \cmidrule(lr){1-1}      \cmidrule(lr){2-4}              \cmidrule(lr){5-7} \cmidrule(lr){8-10} \cmidrule(lr){11-13}
  \input{tables/rgb_auto_colorized}
  \bottomrule
\end{tabular}

%% file: tables/rgb_auto_colorized.tex
nerfacto   &\cellcolor[rgb]{0.5,1,0.5} 18.789 &\cellcolor[rgb]{0.9999811099000832,1,0.9999811099000832}  0.868 &\cellcolor[rgb]{1.0,1,1.0}  0.089     &\cellcolor[rgb]{1.0,1,1.0}      16.147 &\cellcolor[rgb]{0.9981963218310892,1,0.9981963218310892}  0.714 &\cellcolor[rgb]{1.0,1,1.0}  0.229     &\cellcolor[rgb]{0.9999999995732687,1,0.9999999995732687}      11.760 &\cellcolor[rgb]{0.9999999997986778,1,0.9999999997986778}  0.338 &\cellcolor[rgb]{0.71580302812135,1,0.71580302812135}  0.492     &\cellcolor[rgb]{0.9999999999999537,1,0.9999999999999537}      10.371 &\cellcolor[rgb]{1.0,1,1.0}  0.220 &\cellcolor[rgb]{0.9389881340995734,1,0.9389881340995734}  0.560 \\
Ours-Cont  &\cellcolor[rgb]{0.999305291675092,1,0.999305291675092} 18.530 &\cellcolor[rgb]{1.0,1,1.0}  0.861 &\cellcolor[rgb]{0.998046875,1,0.998046875}  0.086     &\cellcolor[rgb]{0.999996969646973,1,0.999996969646973}      16.288 &\cellcolor[rgb]{0.9367361182896108,1,0.9367361182896108}  0.742 &\cellcolor[rgb]{0.9999999999988513,1,0.9999999999988513}  0.227     &\cellcolor[rgb]{0.9999996136278253,1,0.9999996136278253}      12.168 &\cellcolor[rgb]{0.9999918765583887,1,0.9999918765583887}  0.385 &\cellcolor[rgb]{1.0,1,1.0}  0.533     &\cellcolor[rgb]{0.9512348340416366,1,0.9512348340416366}      10.741 &\cellcolor[rgb]{0.999970425639529,1,0.999970425639529}  0.289 &\cellcolor[rgb]{0.9459530059631578,1,0.9459530059631578}  0.562 \\
Ours-RGB   &\cellcolor[rgb]{0.9923468768389357,1,0.9923468768389357} 18.601 &\cellcolor[rgb]{0.9999997852516352,1,0.9999997852516352}  0.865 &\cellcolor[rgb]{0.5,1,0.5}  0.083     &\cellcolor[rgb]{0.5,1,0.5}      16.780 &\cellcolor[rgb]{0.5,1,0.5}  0.765 &\cellcolor[rgb]{0.9999686986888271,1,0.9999686986888271}  0.212     &\cellcolor[rgb]{1.0,1,1.0}      11.456 &\cellcolor[rgb]{1.0,1,1.0}  0.321 &\cellcolor[rgb]{0.9608665245831359,1,0.9608665245831359}  0.501     &\cellcolor[rgb]{0.5,1,0.5}      10.870 &\cellcolor[rgb]{0.999893339703369,1,0.999893339703369}  0.301 &\cellcolor[rgb]{0.5,1,0.5}  0.520 \\
Ours-HS &\cellcolor[rgb]{1.0,1,1.0} 18.327 &\cellcolor[rgb]{0.5,1,0.5}  0.886 &\cellcolor[rgb]{0.5,1,0.5}  0.083     &\cellcolor[rgb]{0.9870313340075252,1,0.9870313340075252}      16.548 &\cellcolor[rgb]{1.0,1,1.0}  0.664 &\cellcolor[rgb]{0.5,1,0.5}  0.172     &\cellcolor[rgb]{0.5,1,0.5}      15.591 &\cellcolor[rgb]{0.5,1,0.5}  0.575 &\cellcolor[rgb]{0.5,1,0.5}  0.489     &\cellcolor[rgb]{1.0,1,1.0}      10.359 &\cellcolor[rgb]{0.5,1,0.5}  0.453 &\cellcolor[rgb]{1.0,1,1.0}  0.693 \\

%% file: tables/rgb_bayspec.tex
\begin{tabular}[htp]{l ccc ccc ccc} \toprule
  \multirow{2}{*}{Method} & \multicolumn{3}{c}{\textit{Anacampseros}} & \multicolumn{3}{c}{\textit{Caladium}} & \multicolumn{3}{c}{Pinecone} \\
           & \PSNR        & \SSIM & \LPIPS & \PSNR           & \SSIM & \LPIPS & \PSNR & \SSIM & \LPIPS \\
  \cmidrule(lr){1-1}      \cmidrule(lr){2-4}              \cmidrule(lr){5-7} \cmidrule(lr){8-10}
  \input{tables/rgb_auto_bayspec_colorized}
  \bottomrule
\end{tabular}

%% file: tables/rgb_auto_bayspec_colorized.tex
nerfacto   &\cellcolor[rgb]{0.9999999999999956,1,0.9999999999999956} 14.413 &\cellcolor[rgb]{0.999999999999282,1,0.999999999999282}  0.626 &\cellcolor[rgb]{0.9999999999999983,1,0.9999999999999983}  0.432     &\cellcolor[rgb]{1.0,1,1.0}      14.190 &\cellcolor[rgb]{0.9999999999975132,1,0.9999999999975132}  0.555 &\cellcolor[rgb]{0.7553035510234112,1,0.7553035510234112}  0.503     &\cellcolor[rgb]{0.9999999999999938,1,0.9999999999999938}      14.120 &\cellcolor[rgb]{1.0,1,1.0}  0.323 &\cellcolor[rgb]{0.8372271785727107,1,0.8372271785727107}  0.464 \\
Ours-Cont  &\cellcolor[rgb]{1.0,1,1.0} 14.171 &\cellcolor[rgb]{0.999999999999282,1,0.999999999999282}  0.626 &\cellcolor[rgb]{1.0,1,1.0}  0.437     &\cellcolor[rgb]{1.0,1,1.0}      14.306 &\cellcolor[rgb]{1.0,1,1.0}  0.543 &\cellcolor[rgb]{0.9507656299504499,1,0.9507656299504499}  0.507     &\cellcolor[rgb]{1.0,1,1.0}      13.868 &\cellcolor[rgb]{0.9999999432897442,1,0.9999999432897442}  0.382 &\cellcolor[rgb]{0.5,1,0.5}  0.414 \\
Ours-RGB   &\cellcolor[rgb]{1.0,1,1.0} 14.321 &\cellcolor[rgb]{1.0,1,1.0}  0.619 &\cellcolor[rgb]{1.0,1,1.0}  0.435     &\cellcolor[rgb]{1.0,1,1.0}      14.249 &\cellcolor[rgb]{0.999999994330388,1,0.999999994330388}  0.569 &\cellcolor[rgb]{0.5,1,0.5}  0.501     &\cellcolor[rgb]{0.9999995398205239,1,0.9999995398205239}      15.412 &\cellcolor[rgb]{0.5,1,0.5}  0.615 &\cellcolor[rgb]{0.7291067427204823,1,0.7291067427204823}  0.442 \\
Ours-HS &\cellcolor[rgb]{0.5,1,0.5} 20.315 &\cellcolor[rgb]{0.5,1,0.5}  0.726 &\cellcolor[rgb]{0.5,1,0.5}  0.297     &\cellcolor[rgb]{0.5,1,0.5}      19.084 &\cellcolor[rgb]{0.5,1,0.5}  0.705 &\cellcolor[rgb]{1.0,1,1.0}  0.530     &\cellcolor[rgb]{0.5,1,0.5}      20.066 &\cellcolor[rgb]{0.8605325940731101,1,0.8605325940731101}  0.580 &\cellcolor[rgb]{1.0,1,1.0}  0.885 \\

%% file: tables/ablation.tex
$\displaystyle
\begin{array}{rcc | ccc ccc ccc}\toprule
  % \rotatebox{-90}{Emittance Network} & \rotatebox{-90}{Transmittance Network} & \rotatebox{-90}{Proposal Network} & \multicolumn{3}{c}{Basil} & \multicolumn{3}{c}{Tools} \\
  \multicolumn{3}{c|}{\text{Network}} & \multicolumn{3}{c}{\text{Rosemary}} & \multicolumn{3}{c}{\text{Basil}} & \multicolumn{3}{c}{\text{Tools}} \\
  \multicolumn{3}{c|}{\text{Archictecure}} & \text{\PSNR} & \text{\SSIM} & \text{\LPIPS}  & \text{\PSNR} & \text{\SSIM} & \text{\LPIPS} & \text{\PSNR} & \text{\SSIM} & \text{\LPIPS} \\
   \cmidrule(lr){1-3} \cmidrule(lr){4-6} \cmidrule(lr){7-9} \cmidrule(lr){10-12}
  %  \text{\hspace*{-6.5em}(HS-NeRF)\hspace*{0.5em}} 
  % \input{tables/ablation_auto_colorized}
  % \\
  % \hline 
  \input{tables/ablation_auto_alt_colorized}
  \\
  \bottomrule
\end{array}
$

%% file: tables/ablation_auto_alt_colorized.tex
\bm{C}_1 & \sigma_0 & P_0      &\cellcolor[rgb]{0.9989300598008057,1,0.9989300598008057} 18.703 &\cellcolor[rgb]{0.9793013022588487,1,0.9793013022588487}  0.883 &\cellcolor[rgb]{0.9961056377100944,1,0.9961056377100944}  0.096     &\cellcolor[rgb]{0.9998852815496433,1,0.9998852815496433}      15.820 &\cellcolor[rgb]{0.9816029333682859,1,0.9816029333682859}  0.756 &\cellcolor[rgb]{0.998559093918138,1,0.998559093918138}  0.234     &\cellcolor[rgb]{0.6805095915200885,1,0.6805095915200885}      15.335 &\cellcolor[rgb]{0.9257498479433259,1,0.9257498479433259}  0.588 &\cellcolor[rgb]{0.9761330572160312,1,0.9761330572160312}  0.553 \\
\bm{C}_1 & \sigma_1 & P_0      &\cellcolor[rgb]{0.9999119988152781,1,0.9999119988152781} 18.432 &\cellcolor[rgb]{0.99951171875,1,0.99951171875}  0.873 &\cellcolor[rgb]{0.9999081321873169,1,0.9999081321873169}  0.106     &\cellcolor[rgb]{0.9996692079367389,1,0.9996692079367389}      15.882 &\cellcolor[rgb]{0.9718432426452638,1,0.9718432426452638}  0.760 &\cellcolor[rgb]{0.9997672856233594,1,0.9997672856233594}  0.247     &\cellcolor[rgb]{0.9088677398998146,1,0.9088677398998146}      14.375 &\cellcolor[rgb]{0.9999940930116535,1,0.9999940930116535}  0.431 &\cellcolor[rgb]{0.9999999999576177,1,0.9999999999576177}  0.709 \\
(ours)~\bm{C}_{\phantom{~}} & \sigma_0 & P_0  &\cellcolor[rgb]{0.9999725435704786,1,0.9999725435704786} 18.327 &\cellcolor[rgb]{0.9492864884986142,1,0.9492864884986142}  0.886 &\cellcolor[rgb]{0.8822226697131839,1,0.8822226697131839}  0.083     &\cellcolor[rgb]{0.5,1,0.5}      16.548 &\cellcolor[rgb]{1.0,1,1.0}  0.664 &\cellcolor[rgb]{0.5,1,0.5}  0.172     &\cellcolor[rgb]{0.564599133229063,1,0.564599133229063}      15.591 &\cellcolor[rgb]{0.9558160664101377,1,0.9558160664101377}  0.575 &\cellcolor[rgb]{0.5,1,0.5}  0.489 \\
\bm{C}_{\phantom{~}} & \sigma & P_0   &\cellcolor[rgb]{1.0,1,1.0} 17.477 &\cellcolor[rgb]{0.9999988617027151,1,0.9999988617027151}  0.863 &\cellcolor[rgb]{0.9782843649597519,1,0.9782843649597519}  0.090     &\cellcolor[rgb]{0.5589685202831018,1,0.5589685202831018}      16.532 &\cellcolor[rgb]{0.5,1,0.5}  0.792 &\cellcolor[rgb]{0.9990265754804508,1,0.9990265754804508}  0.237     &\cellcolor[rgb]{1.0,1,1.0}       7.192 &\cellcolor[rgb]{1.0,1,1.0}  0.331 &\cellcolor[rgb]{1.0,1,1.0}  0.733 \\
\bm{C}_2 & \sigma_2 & P_0      &\cellcolor[rgb]{0.5,1,0.5} 19.744 &\cellcolor[rgb]{0.5,1,0.5}  0.895 &\cellcolor[rgb]{0.5,1,0.5}  0.076     &\cellcolor[rgb]{0.8620268952777661,1,0.8620268952777661}      16.393 &\cellcolor[rgb]{0.868462211918086,1,0.868462211918086}  0.776 &\cellcolor[rgb]{0.9718432426452637,1,0.9718432426452637}  0.207     &\cellcolor[rgb]{0.5,1,0.5}      15.708 &\cellcolor[rgb]{0.5,1,0.5}  0.642 &\cellcolor[rgb]{0.9900020448758763,1,0.9900020448758763}  0.568 \\
\bm{C}_{\phantom{~}} & \sigma & P_\lambda    &\cellcolor[rgb]{1.0,1,1.0} 17.494 &\cellcolor[rgb]{1.0,1,1.0}  0.851 &\cellcolor[rgb]{1.0,1,1.0}  0.128     &\cellcolor[rgb]{1.0,1,1.0}      15.265 &\cellcolor[rgb]{0.9999955591079015,1,0.9999955591079015}  0.704 &\cellcolor[rgb]{1.0,1,1.0}  0.312     &\cellcolor[rgb]{0.9841851049096417,1,0.9841851049096417}      13.221 &\cellcolor[rgb]{0.9999998751308253,1,0.9999998751308253}  0.399 &\cellcolor[rgb]{0.9999981117819664,1,0.9999981117819664}  0.663 

%% file: tables/ablation_bayspec.tex
$\displaystyle
\begin{array}{rcc | ccc ccc ccc}\toprule
  % \rotatebox{-90}{Emittance Network} & \rotatebox{-90}{Transmittance Network} & \rotatebox{-90}{Proposal Network} & \multicolumn{3}{c}{Basil} & \multicolumn{3}{c}{Tools} \\
  \multicolumn{3}{c|}{\text{Network}} & \multicolumn{3}{c}{\text{\textit{Anacampseros}}} & \multicolumn{3}{c}{\text{\textit{Caladium}}} & \multicolumn{3}{c}{\text{Pinecone}} \\
  \multicolumn{3}{c|}{\text{Archictecure}} & \text{\PSNR} & \text{\SSIM} & \text{\LPIPS}  & \text{\PSNR} & \text{\SSIM} & \text{\LPIPS} & \text{\PSNR} & \text{\SSIM} & \text{\LPIPS} \\
   \cmidrule(lr){1-3} \cmidrule(lr){4-6} \cmidrule(lr){7-9} \cmidrule(lr){10-12}
  %  \text{\hspace*{-6.5em}(HS-NeRF)\hspace*{0.5em}} 
  \input{tables/ablation_auto_bayspec_colorized}
  \\
  \bottomrule
\end{array}
$

%% file: tables/ablation_auto_bayspec_colorized.tex
\bm{C}_1 & \sigma_0 & P_0      &\cellcolor[rgb]{0.8729040288044295,1,0.8729040288044295} 20.279 &\cellcolor[rgb]{0.5,1,0.5}  0.734 &\cellcolor[rgb]{0.5,1,0.5}  0.295     &\cellcolor[rgb]{0.9257359216835326,1,0.9257359216835326}      19.084 &\cellcolor[rgb]{0.5,1,0.5}  0.705 &\cellcolor[rgb]{0.9973043972155617,1,0.9973043972155617}  0.530     &\cellcolor[rgb]{0.5513230257097693,1,0.5513230257097693}      20.102 &\cellcolor[rgb]{0.6486279830649155,1,0.6486279830649155}  0.729 &\cellcolor[rgb]{0.9449728235767153,1,0.9449728235767153}  0.399 \\
\bm{C}_1 & \sigma_1 & P_0      &\cellcolor[rgb]{0.8993767728185065,1,0.8993767728185065} 20.128 &\cellcolor[rgb]{0.731153059388185,1,0.731153059388185}  0.729 &\cellcolor[rgb]{0.8963597524286337,1,0.8963597524286337}  0.305     &\cellcolor[rgb]{0.5,1,0.5}      19.933 &\cellcolor[rgb]{0.9474644525659037,1,0.9474644525659037}  0.678 &\cellcolor[rgb]{0.5,1,0.5}  0.495     &\cellcolor[rgb]{0.7162253002729333,1,0.7162253002729333}      19.857 &\cellcolor[rgb]{0.6817379165683841,1,0.6817379165683841}  0.726 &\cellcolor[rgb]{0.5,1,0.5}  0.372 \\
(ours)~\bm{C}_{\phantom{~}} & \sigma_0 & P_0  &\cellcolor[rgb]{0.8657594741527114,1,0.8657594741527114} 20.315 &\cellcolor[rgb]{0.8192044753625585,1,0.8192044753625585}  0.726 &\cellcolor[rgb]{0.6262201659534103,1,0.6262201659534103}  0.297     &\cellcolor[rgb]{0.8300020341976766,1,0.8300020341976766}      19.428 &\cellcolor[rgb]{0.8802898040092013,1,0.8802898040092013}  0.687 &\cellcolor[rgb]{0.9895747318685465,1,0.9895747318685465}  0.523     &\cellcolor[rgb]{0.5,1,0.5}      20.162 &\cellcolor[rgb]{0.5,1,0.5}  0.740 &\cellcolor[rgb]{0.9797830363557996,1,0.9797830363557996}  0.409 \\
\bm{C}_{\phantom{~}} & \sigma & P_0   &\cellcolor[rgb]{0.904471985942854,1,0.904471985942854} 20.095 &\cellcolor[rgb]{0.99464646858295,1,0.99464646858295}  0.705 &\cellcolor[rgb]{0.9955908046679508,1,0.9955908046679508}  0.320     &\cellcolor[rgb]{0.9486281715365203,1,0.9486281715365203}      18.942 &\cellcolor[rgb]{0.9970128813623071,1,0.9970128813623071}  0.653 &\cellcolor[rgb]{0.9551732552758285,1,0.9551732552758285}  0.514     &\cellcolor[rgb]{0.8310784482065837,1,0.8310784482065837}      19.596 &\cellcolor[rgb]{0.6486279830649155,1,0.6486279830649155}  0.729 &\cellcolor[rgb]{0.9797830363557996,1,0.9797830363557996}  0.409 \\
\bm{C}_2 & \sigma_2 & P_0      &\cellcolor[rgb]{0.5,1,0.5} 21.259 &\cellcolor[rgb]{0.9827021154333857,1,0.9827021154333857}  0.711 &\cellcolor[rgb]{0.9943158488905341,1,0.9943158488905341}  0.319     &\cellcolor[rgb]{0.9418538319166285,1,0.9418538319166285}      18.989 &\cellcolor[rgb]{1.0,1,1.0}  0.595 &\cellcolor[rgb]{1.0,1,1.0}  0.568     &\cellcolor[rgb]{0.7777213941179423,1,0.7777213941179423}      19.732 &\cellcolor[rgb]{0.9999688715641195,1,0.9999688715641195}  0.561 &\cellcolor[rgb]{0.9992111655703528,1,0.9992111655703528}  0.434 \\
\bm{C}_{\phantom{~}} & \sigma & P_\lambda    &\cellcolor[rgb]{1.0,1,1.0} 15.031 &\cellcolor[rgb]{1.0,1,1.0}  0.667 &\cellcolor[rgb]{1.0,1,1.0}  0.351     &\cellcolor[rgb]{1.0,1,1.0}      15.930 &\cellcolor[rgb]{0.999980106001826,1,0.999980106001826}  0.626 &\cellcolor[rgb]{0.9999997334141514,1,0.9999997334141514}  0.556     &\cellcolor[rgb]{1.0,1,1.0}      15.700 &\cellcolor[rgb]{1.0,1,1.0}  0.485 &\cellcolor[rgb]{1.0,1,1.0}  0.484 

%% file: figs/abblation_grid.tex
\begin{figure}
  \centering
  \setlength{\tabcolsep}{2pt}
  \scriptsize
  \begin{tabular}[t]{c|cccccccc}
    & \parbox[b]{0.112\linewidth}{\centering Ground\\Truth} & $\bm{C}_1\sigma_0P_0$ & $\bm{C}_1\sigma_1P_0$ & \parbox[b]{0.112\linewidth}{\centering (ours)\\$\bm{C}\sigma_0P_0$} & $\bm{C}\sigma P_0$ & $\bm{C}_2\sigma_2P_0$ & $\bm{C}\sigma P_\lambda$ \\
    \hline \\[-0.75em]
    \parbox[b]{0.112\linewidth}{\centering Channel \\ 15 \\ (447nm) \\[0.25em]}
    & \includegraphics[width=0.112\linewidth]{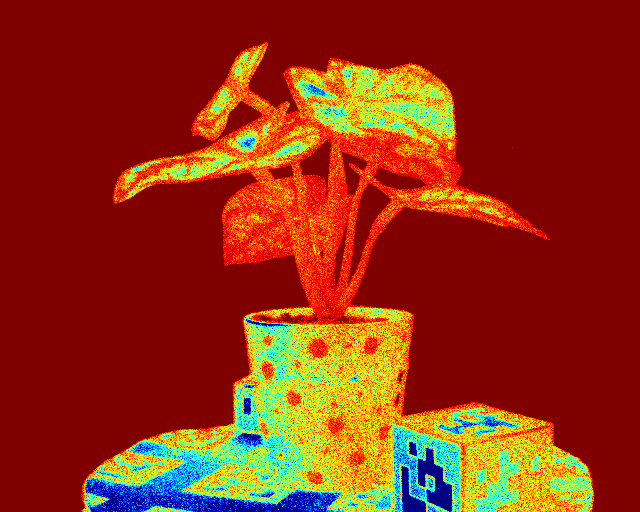} & \includegraphics[width=0.112\linewidth]{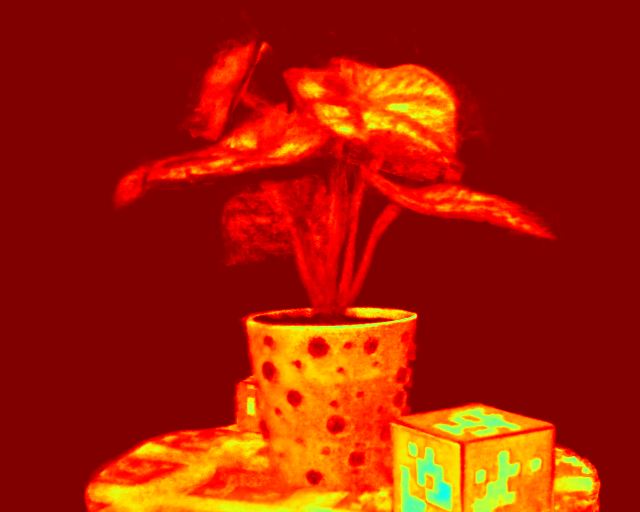} & \includegraphics[width=0.112\linewidth]{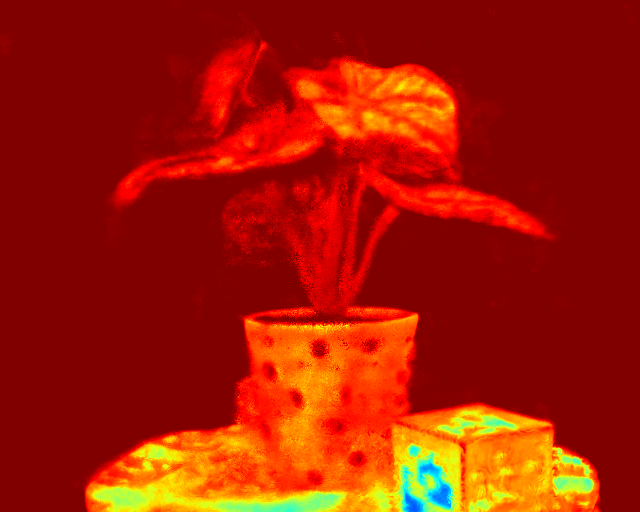} & \includegraphics[width=0.112\linewidth]{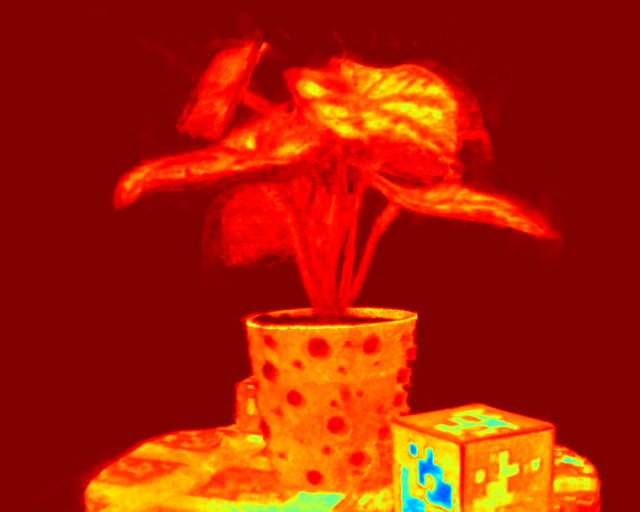} & \includegraphics[width=0.112\linewidth]{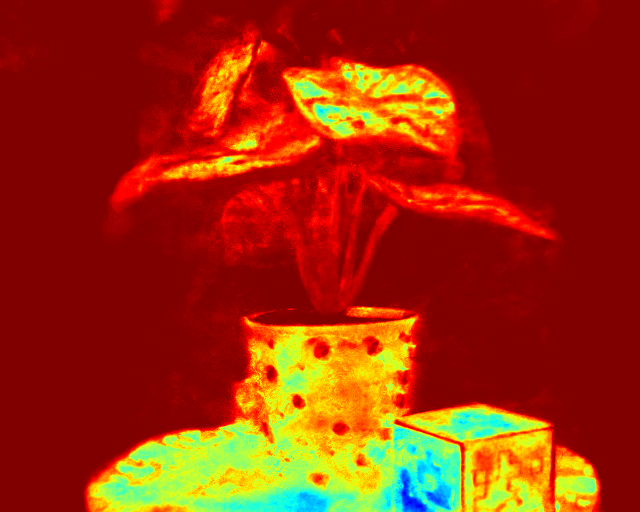} & \includegraphics[width=0.112\linewidth]{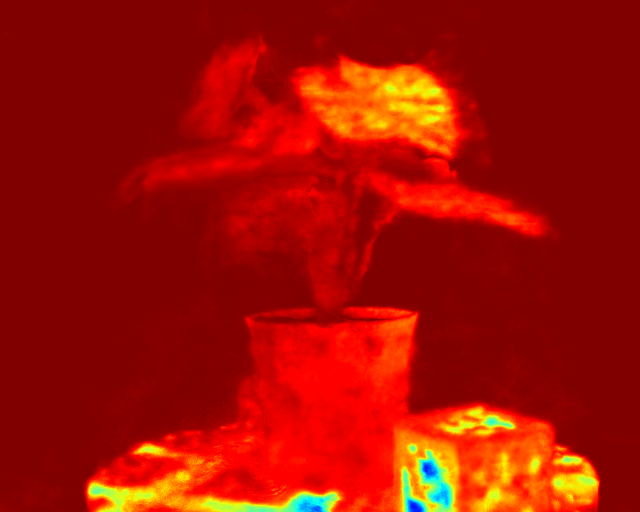} & \includegraphics[width=0.112\linewidth]{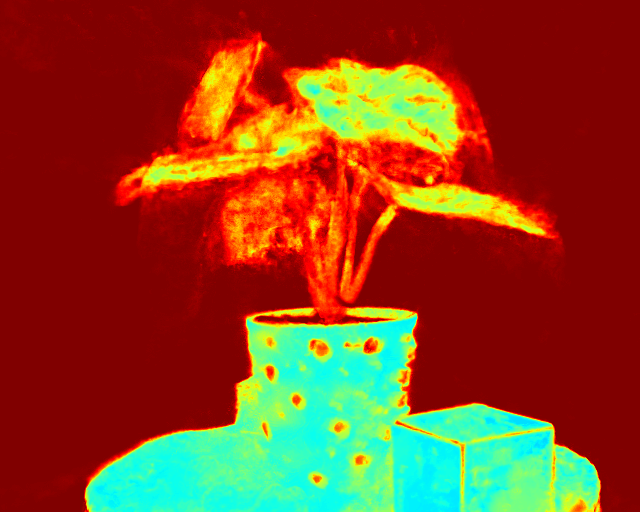} \\
    \parbox[b]{0.112\linewidth}{\centering Channel \\ 55 \\ (654nm) \\[0.25em]}
    & \includegraphics[width=0.112\linewidth]{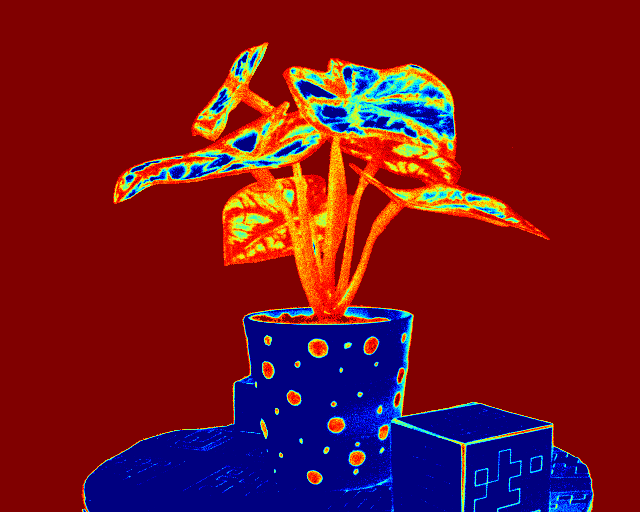} & \includegraphics[width=0.112\linewidth]{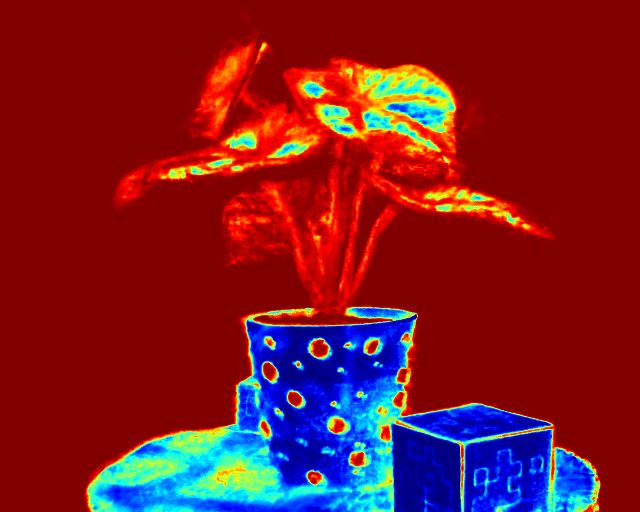} & \includegraphics[width=0.112\linewidth]{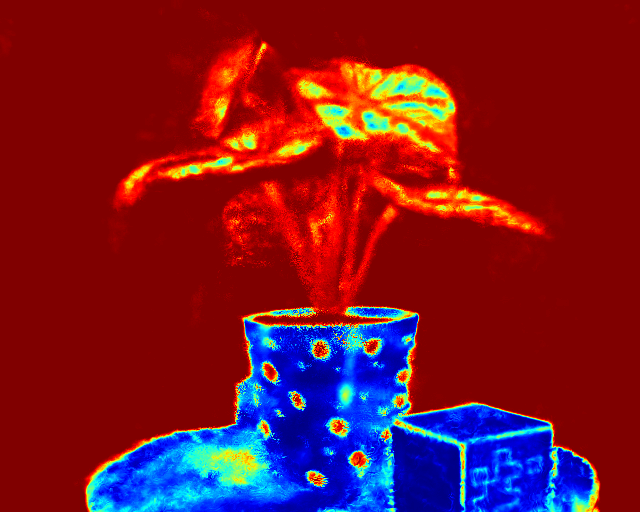} & \includegraphics[width=0.112\linewidth]{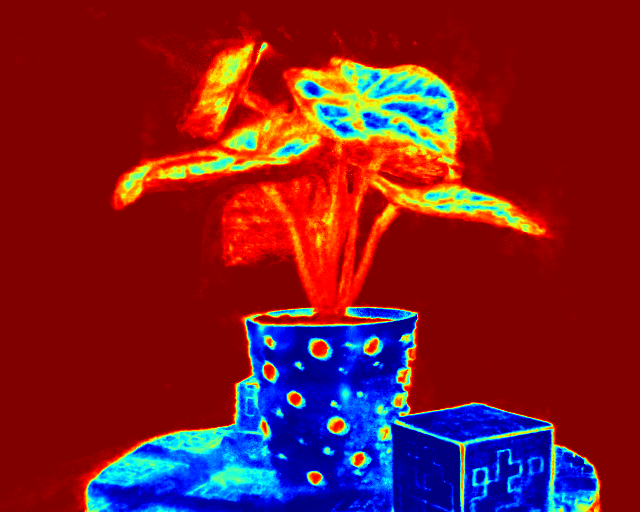} & \includegraphics[width=0.112\linewidth]{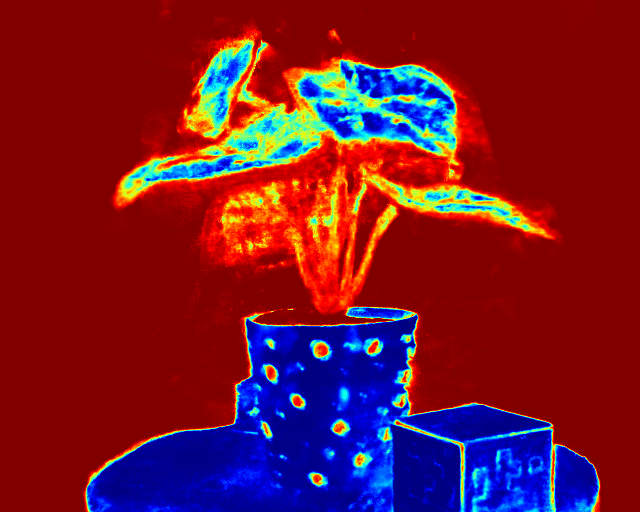} & \includegraphics[width=0.112\linewidth]{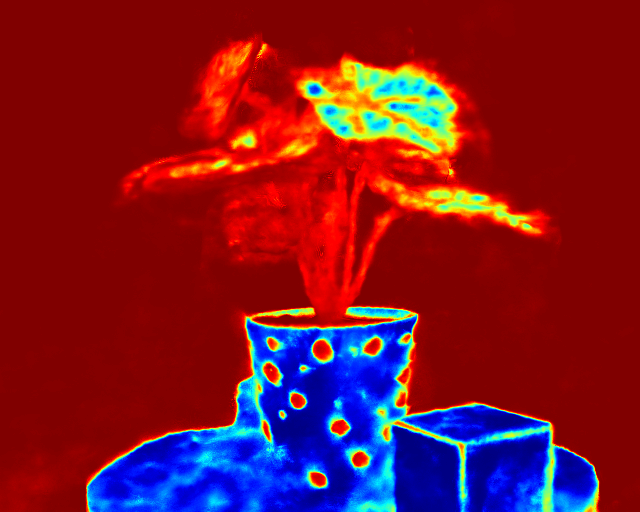} & \includegraphics[width=0.112\linewidth]{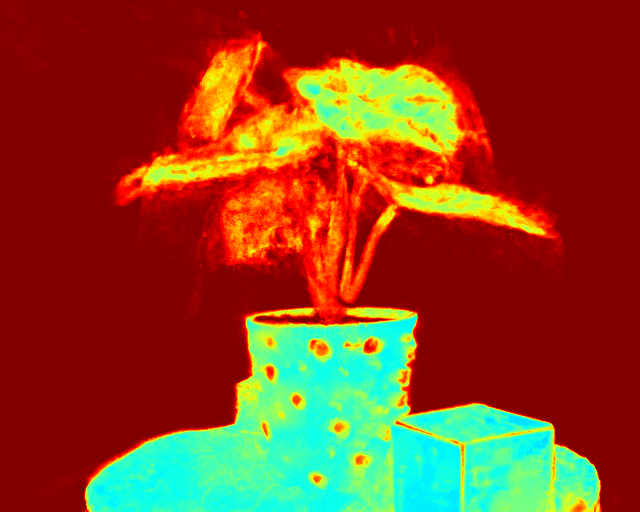} \\
    \parbox[b]{0.112\linewidth}{\centering Channel \\ 95 \\ (868nm) \\[0.25em]}
    & \includegraphics[width=0.112\linewidth]{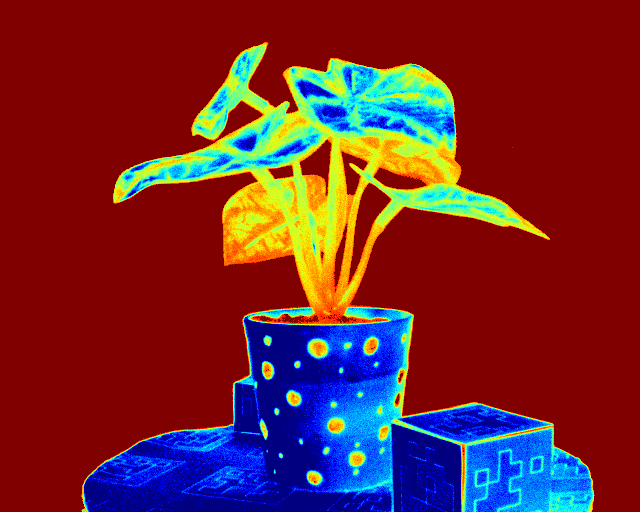} & \includegraphics[width=0.112\linewidth]{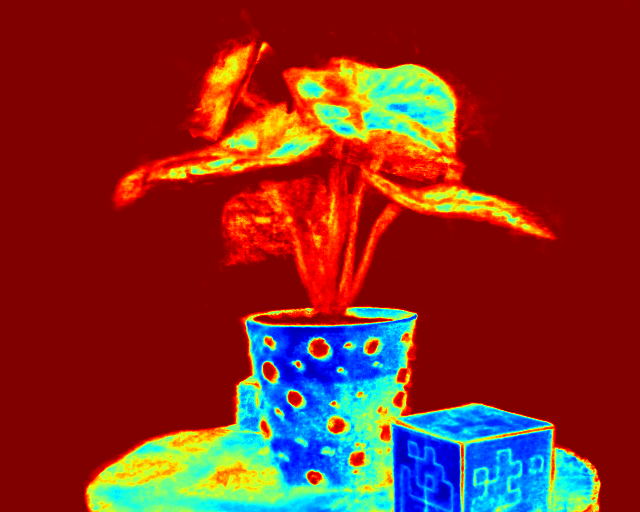} & \includegraphics[width=0.112\linewidth]{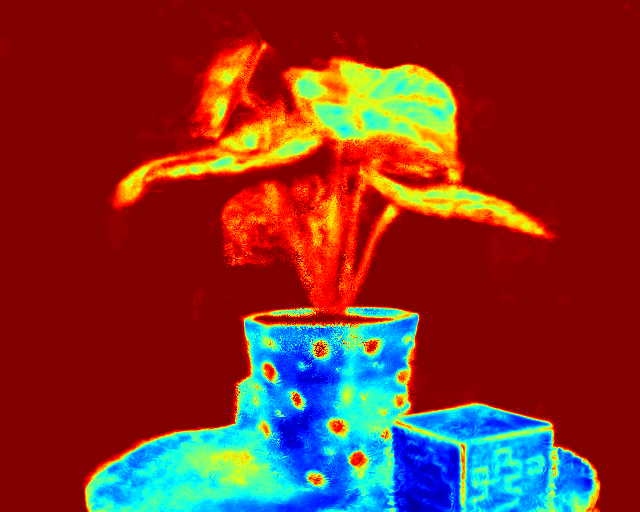} & \includegraphics[width=0.112\linewidth]{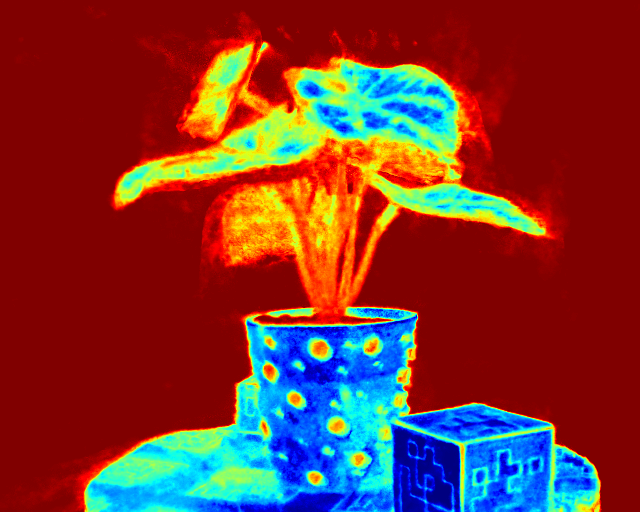} & \includegraphics[width=0.112\linewidth]{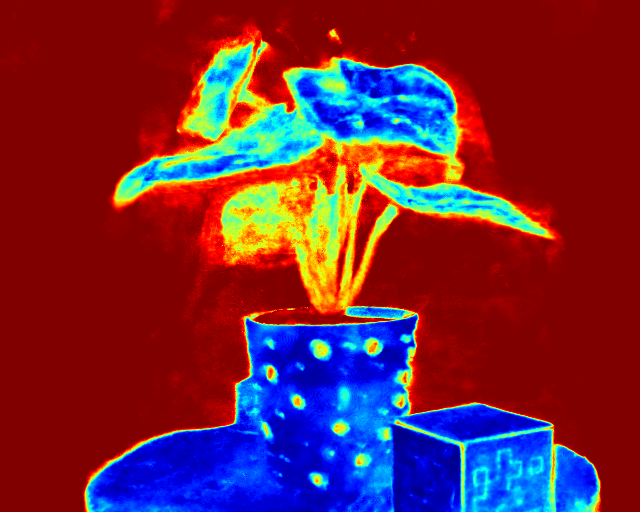} & \includegraphics[width=0.112\linewidth]{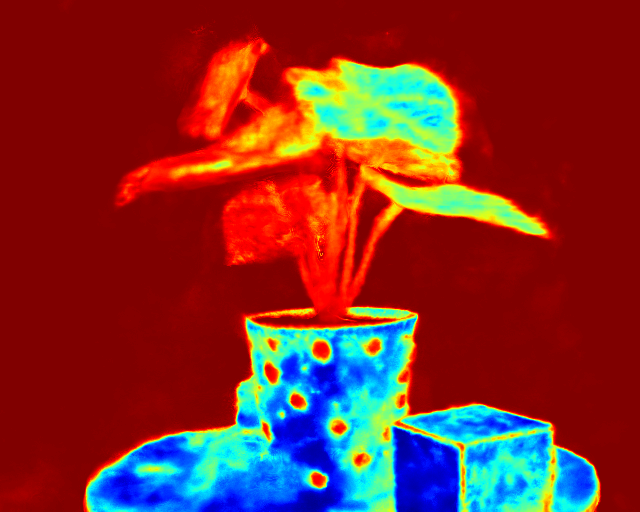} & \includegraphics[width=0.112\linewidth]{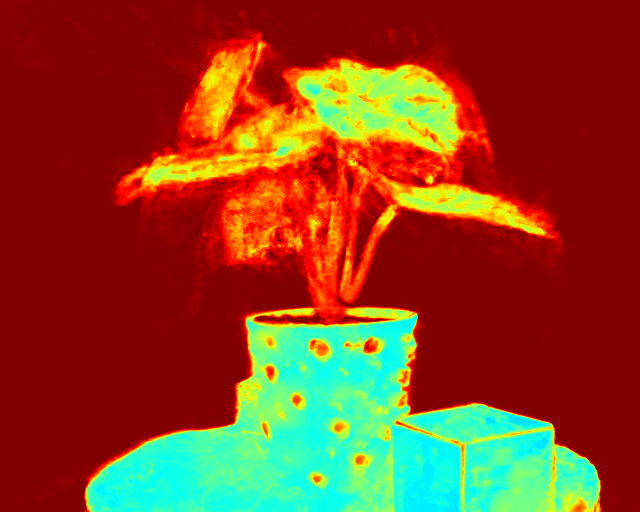} \\
  \end{tabular}

  \caption{Comparing hyperspectral images (\textit{Caladium} dataset, validation image) rendered by different networks, we see that most methods perform similarly with the exception of the wavelength-dependent proposal network.  % Note that, unlike Figure \ref{fig:leave_out_wavelengths}, these wavelengths appear in the training sets but these images do not.
  }
  \label{fig:ablation}
\end{figure}

%% file: 4_results_superresolution.tex
\begin{table}
  \centering
  \caption{Having 128 channels for each image allows us to withhold wavelengths from the training set and force the network to interpolate.  The relatively small drop in performance when withholding even the vast majority of the wavelengths supports the claim that continuous radiance and transmission spectra are well suited for HS-NeRF.}
  \scriptsize
  Basil Dataset
  \input{tables/wavelength}\\[0.75em]
  \textit{Anacampseros} Dataset
  \input{tables/wavelength_bayspec}
  \label{tab:leave_out_wavelengths}
\end{table}

\begin{figure}
  \centering
  \setlength{\tabcolsep}{2pt}
  \small
  \scalebox{1}{
  \begin{tabular}[t]{c|ccccc}
    \# Wavelengths & Channel 15 & Channel 35 & Channel 55 & Channel 75 & Channel 95 \\
    in Train Set    &  (447nm) &  (550nm) &  (654nm) &  (760nm) &  (868nm) \\[0.2em]
    \hline 
    \parbox[b][0.35in][t]{0.4in}{\centering Ground\\Truth}    &  \includegraphics[width=0.15\linewidth]{figs/eval_images/GROUND_TRUTH/jet_15.png} &  \includegraphics[width=0.15\linewidth]{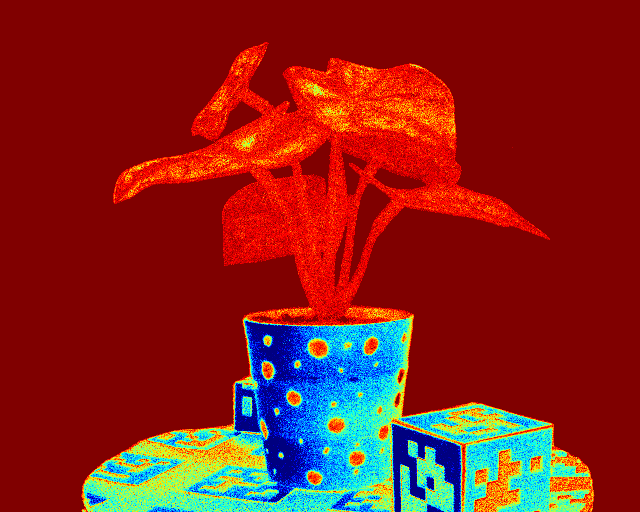} &  \includegraphics[width=0.15\linewidth]{figs/eval_images/GROUND_TRUTH/jet_55.png} &  \includegraphics[width=0.15\linewidth]{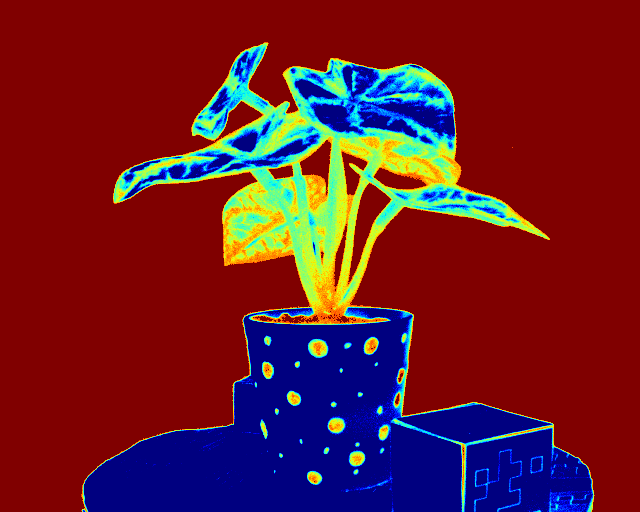} &  \includegraphics[width=0.15\linewidth]{figs/eval_images/GROUND_TRUTH/jet_95.png} \\
    \parbox[b][0.35in][t]{0.4in}{128         }    &  \includegraphics[width=0.15\linewidth]{figs/eval_images/every_1/jet_15.png} &  \includegraphics[width=0.15\linewidth]{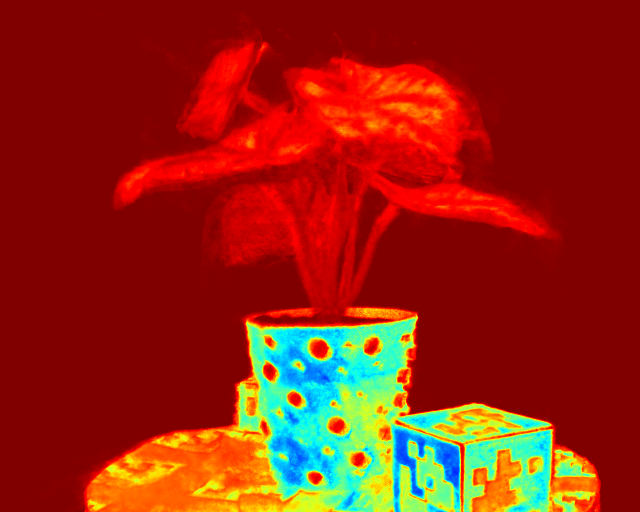} &  \includegraphics[width=0.15\linewidth]{figs/eval_images/every_1/jet_55.png} &  \includegraphics[width=0.15\linewidth]{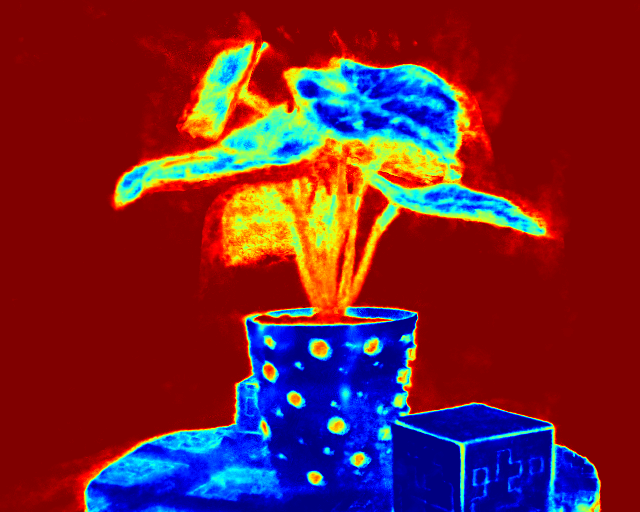} &  \includegraphics[width=0.15\linewidth]{figs/eval_images/every_1/jet_95.png} \\
    \parbox[b][0.35in][t]{0.4in}{64          }    &  \includegraphics[width=0.15\linewidth]{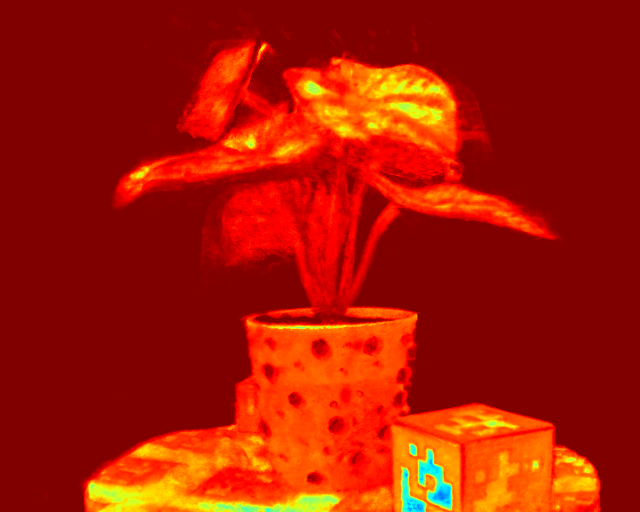} &  \includegraphics[width=0.15\linewidth]{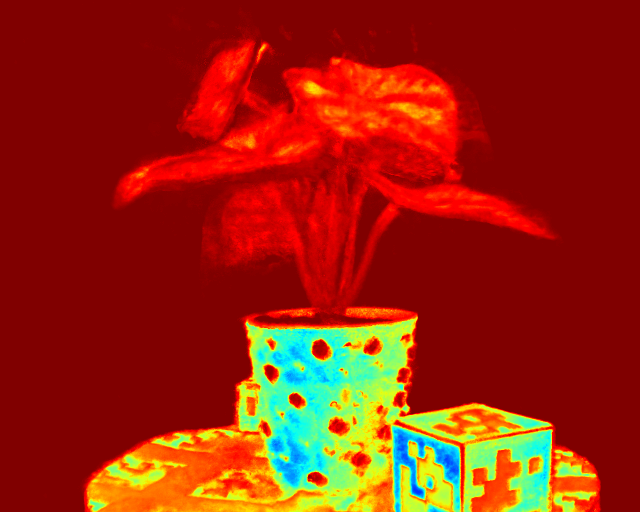} &  \includegraphics[width=0.15\linewidth]{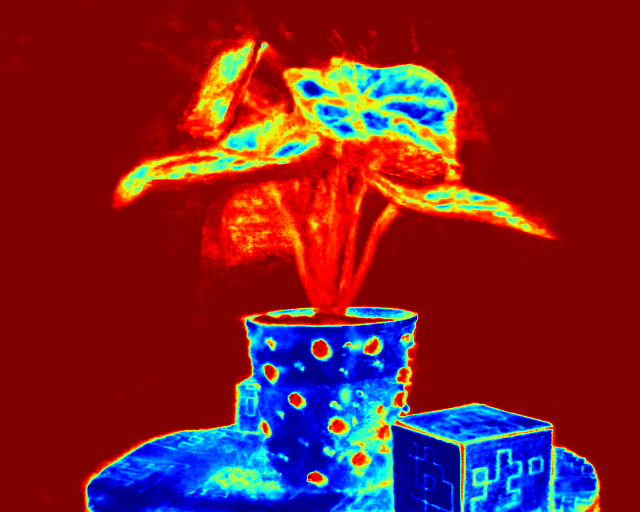} &  \includegraphics[width=0.15\linewidth]{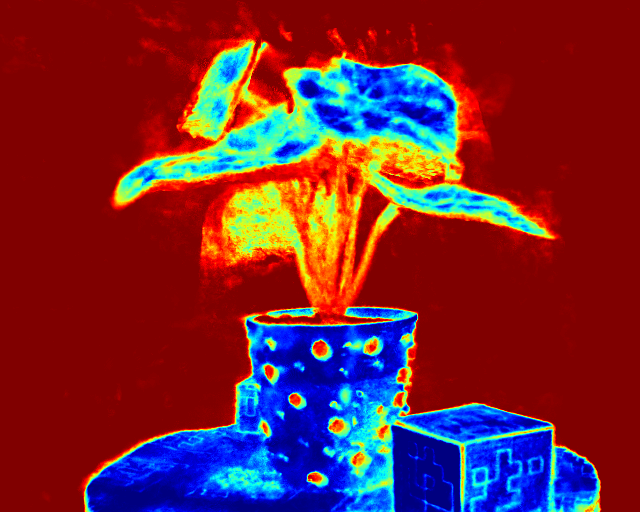} &  \includegraphics[width=0.15\linewidth]{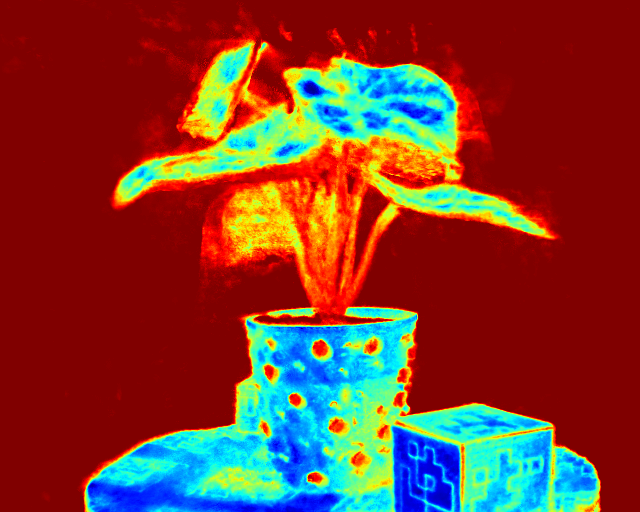} \\
    \parbox[b][0.35in][t]{0.4in}{32          }    &  \includegraphics[width=0.15\linewidth]{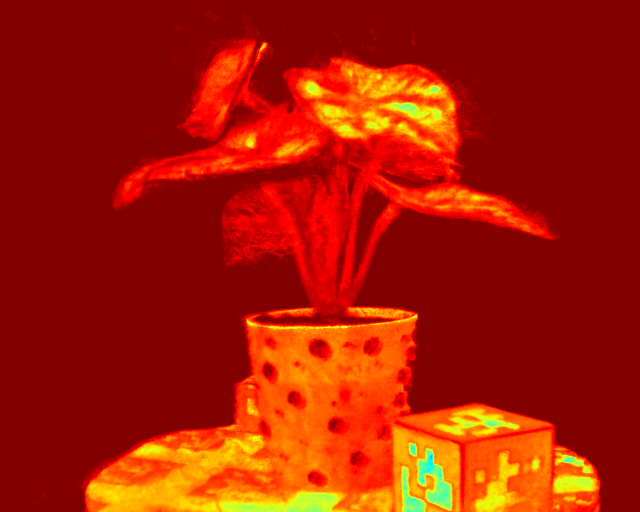} &  \includegraphics[width=0.15\linewidth]{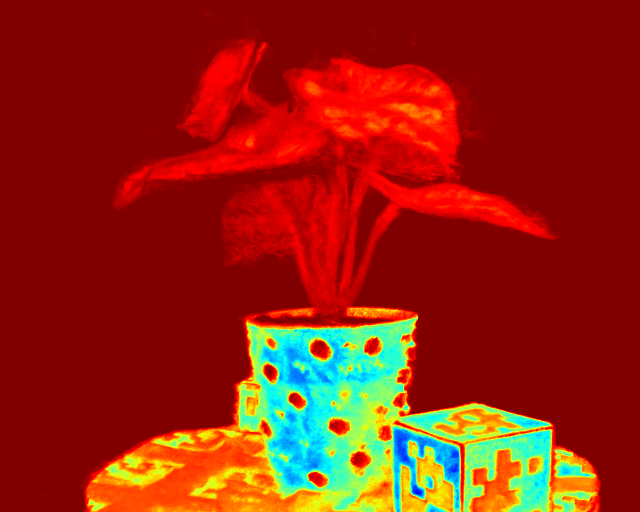} &  \includegraphics[width=0.15\linewidth]{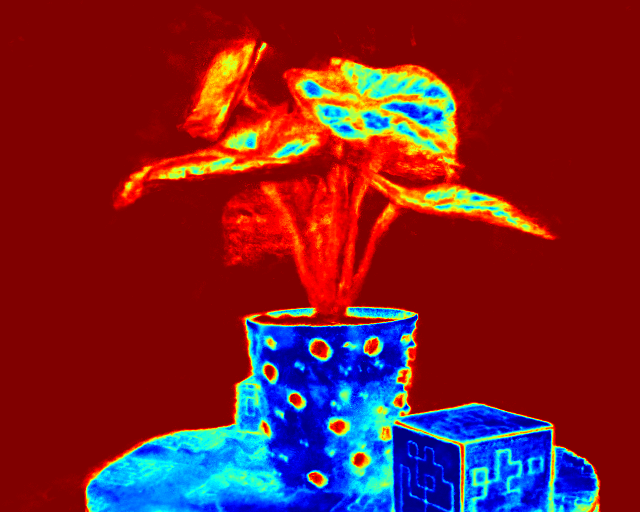} &  \includegraphics[width=0.15\linewidth]{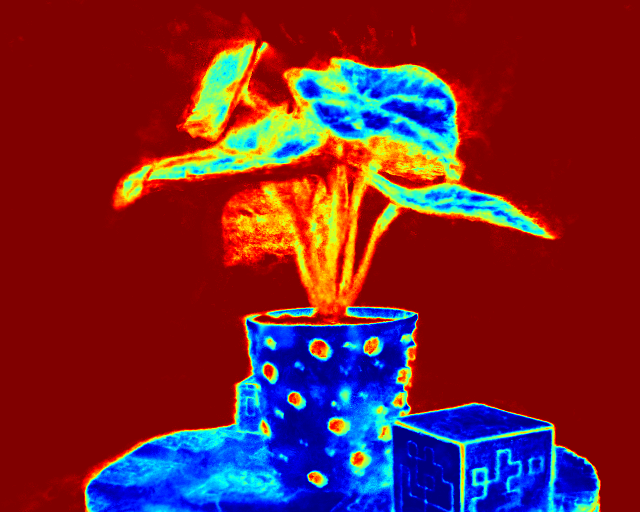} &  \includegraphics[width=0.15\linewidth]{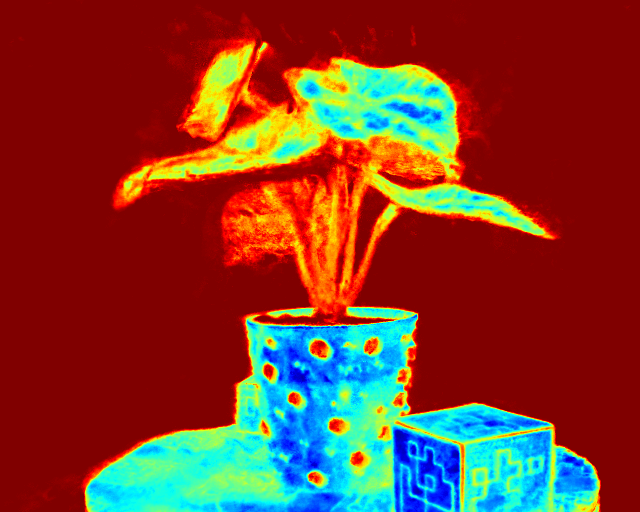} \\
    \parbox[b][0.35in][t]{0.4in}{16          }    &  \includegraphics[width=0.15\linewidth]{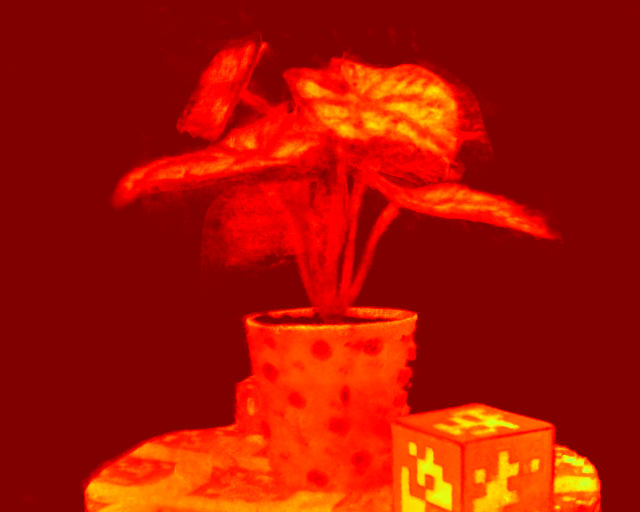} &  \includegraphics[width=0.15\linewidth]{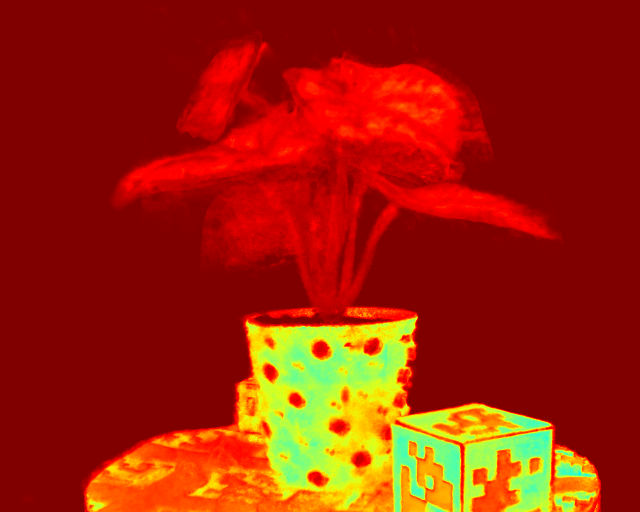} &  \includegraphics[width=0.15\linewidth]{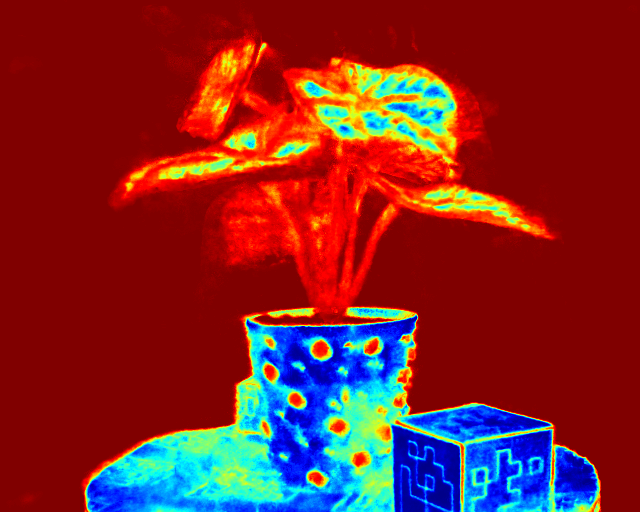} &  \includegraphics[width=0.15\linewidth]{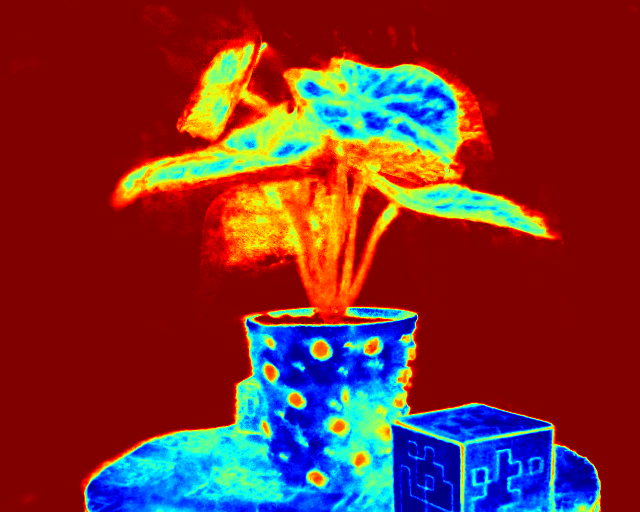} &  \includegraphics[width=0.15\linewidth]{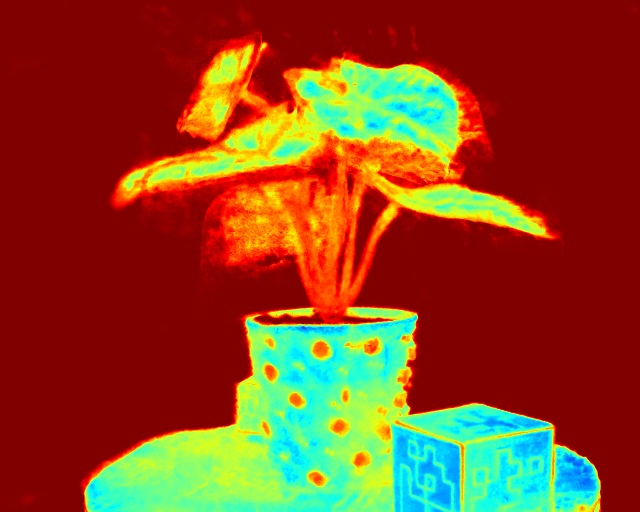} \\
  \end{tabular}
  }
  \caption{We visually observe the ability of HS-NeRF to interpolate wavelengths unseen in the training set.  None of the wavelengths from this image were used in training (except 128 channel case), and none of the images used in training used these wavelengths (corresponds to Both Unseen in Table \ref{tab:leave_out_wavelengths}).  We color the images using the ``jet'' colormap available in matplotlib and matlab for easier perception.}
  \label{fig:leave_out_wavelengths}
\end{figure}

% \subsection{HS-NeRF Wavelength Generalization} \label{ssec:hypernerf_wavelength_validation}
We demonstrate hyperspectral super-resolution on our 8 image sets by witholding entire images and entire wavelengths from the training sets and evaluating the hyperspectral image predictions.
% In particular, we seek to demonstrate 2 things: (1) that NeRF methods generalize well to hyperspectral data and (2) that we can suitably learn continuous representations for spectra.
To characterize the accuracy vs super-resolution factor, we train the same NeRF architecture 4 separate times: first using the full 128 wavelengths, then with only 64, 32, and 16 evenly sampled wavelengths.  During evaluation, the networks must generalize to both unseen images \emph{and} unseen wavelengths.

Table \ref{tab:leave_out_wavelengths} and Figure \ref{fig:leave_out_wavelengths} illustrate that the same network architecture can incorporate arbitrary wavelength supervision: 
increasing or decreasing the number of wavelengths used during training has a minimal effect on evaluation accuracy.  From this we can deduce that \emph{continuous} representations of radiance spectra can successfully allow generalizing NeRF to arbitrary wavelengths.

%% file: tables/wavelength.tex
\begin{tabular}[pos]{c ccc ccc cc cc} \toprule
  \# of Wavelengths & \multicolumn{3}{c}{Train Set} & \multicolumn{3}{c}{Unseen Images} & \multicolumn{2}{c}{Unseen Wavelengths} & \multicolumn{2}{c}{Both Unseen} \\[0.4em]
  in Train Set      & \PSNR        & \SSIM & \LPIPS & \PSNR           & \SSIM & \LPIPS & \PSNR & \SSIM  & \PSNR & \SSIM  \\
  \cmidrule(lr){1-1}      \cmidrule(lr){2-4}              \cmidrule(lr){5-7} \cmidrule(lr){8-9} \cmidrule(lr){10-11}
   \input{tables/wavelength_auto_colorized}
  \bottomrule
\end{tabular}

%% file: tables/wavelength_auto_colorized.tex
128        &\cellcolor[rgb]{0.5,1,0.5} 20.378 &\cellcolor[rgb]{0.5,1,0.5}  0.848 &\cellcolor[rgb]{0.9015627978296388,1,0.9015627978296388}  0.255 &\cellcolor[rgb]{0.999896856099743,1,0.999896856099743} 16.493 &\cellcolor[rgb]{0.9959157977845854,1,0.9959157977845854}  0.798 &\cellcolor[rgb]{0.99999704755,1,0.99999704755}  0.288 & N/A & N/A & N/A & N/A \\
64         &\cellcolor[rgb]{0.7616464953959546,1,0.7616464953959546} 19.893 &\cellcolor[rgb]{0.7546108509882288,1,0.7546108509882288}  0.839 &\cellcolor[rgb]{0.5,1,0.5}  0.246 &\cellcolor[rgb]{0.99988134992315,1,0.99988134992315} 16.534 &\cellcolor[rgb]{0.9991782344040058,1,0.9991782344040058}  0.786 &\cellcolor[rgb]{0.9996521135772666,1,0.9996521135772666}  0.277 &\cellcolor[rgb]{0.715771915115113,1,0.715771915115113} 20.005 &\cellcolor[rgb]{0.8385211435321543,1,0.8385211435321543}  0.834 &\cellcolor[rgb]{0.9998248952123672,1,0.9998248952123672} 16.651 &\cellcolor[rgb]{0.9995477374634228,1,0.9995477374634228}  0.782 \\
32         &\cellcolor[rgb]{0.8829604471640253,1,0.8829604471640253} 19.460 &\cellcolor[rgb]{0.9274738535414557,1,0.9274738535414557}  0.825 &\cellcolor[rgb]{0.9858762375500001,1,0.9858762375500001}  0.264 &\cellcolor[rgb]{0.9999601925037106,1,0.9999601925037106} 16.229 &\cellcolor[rgb]{0.9996126917146082,1,0.9996126917146082}  0.781 &\cellcolor[rgb]{0.9999999917309141,1,0.9999999917309141}  0.296 &\cellcolor[rgb]{0.8855250544063498,1,0.8855250544063498} 19.447 &\cellcolor[rgb]{0.9548529586347493,1,0.9548529586347493}  0.820 &\cellcolor[rgb]{0.9999556025762522,1,0.9999556025762522} 16.258 &\cellcolor[rgb]{0.9997968721126531,1,0.9997968721126531}  0.777 \\
16         &\cellcolor[rgb]{0.999999997459185,1,0.999999997459185} 14.592 &\cellcolor[rgb]{0.9999942621399442,1,0.9999942621399442}  0.759 &\cellcolor[rgb]{0.9982929483137318,1,0.9982929483137318}  0.272 &\cellcolor[rgb]{1.0,1,1.0} 13.586 &\cellcolor[rgb]{1.0,1,1.0}  0.717 &\cellcolor[rgb]{1.0,1,1.0}  0.306 &\cellcolor[rgb]{0.9999999952920591,1,0.9999999952920591} 14.656 &\cellcolor[rgb]{0.9999942621399442,1,0.9999942621399442}  0.759 &\cellcolor[rgb]{1.0,1,1.0} 13.641 &\cellcolor[rgb]{1.0,1,1.0}  0.717 \\

%% file: tables/wavelength_bayspec.tex
\begin{tabular}[pos]{c ccc ccc cc cc} \toprule
  \# of Wavelengths & \multicolumn{3}{c}{Train Set} & \multicolumn{3}{c}{Unseen Images} & \multicolumn{2}{c}{Unseen Wavelengths} & \multicolumn{2}{c}{Both Unseen} \\[0.4em]
  in Train Set      & \PSNR        & \SSIM & \LPIPS & \PSNR           & \SSIM & \LPIPS & \PSNR & \SSIM  & \PSNR & \SSIM  \\
  \cmidrule(lr){1-1}      \cmidrule(lr){2-4}              \cmidrule(lr){5-7} \cmidrule(lr){8-9} \cmidrule(lr){10-11}
   \input{tables/wavelength_auto_bayspec_colorized}
  \bottomrule
\end{tabular}

%% file: tables/wavelength_auto_bayspec_colorized.tex
128        &\cellcolor[rgb]{0.9009277240713697,1,0.9009277240713697} 20.550 &\cellcolor[rgb]{0.9786558295198415,1,0.9786558295198415}  0.730 &\cellcolor[rgb]{0.5,1,0.5}  0.294 &\cellcolor[rgb]{0.9723637950787285,1,0.9723637950787285} 20.315 &\cellcolor[rgb]{0.9999658493272318,1,0.9999658493272318}  0.726 &\cellcolor[rgb]{0.5319894255610764,1,0.5319894255610764}  0.297 & N/A & N/A & N/A & N/A \\
64         &\cellcolor[rgb]{0.5478323939129007,1,0.5478323939129007} 20.947 &\cellcolor[rgb]{0.6584932723174647,1,0.6584932723174647}  0.735 &\cellcolor[rgb]{0.9999707823302408,1,0.9999707823302408}  0.461 &\cellcolor[rgb]{0.8107817780028576,1,0.8107817780028576} 20.701 &\cellcolor[rgb]{0.9180042346834233,1,0.9180042346834233}  0.732 &\cellcolor[rgb]{0.9999531989425561,1,0.9999531989425561}  0.459 &\cellcolor[rgb]{0.5,1,0.5} 20.979 &\cellcolor[rgb]{0.5,1,0.5}  0.736 &\cellcolor[rgb]{0.7864845968605131,1,0.7864845968605131} 20.732 &\cellcolor[rgb]{0.8601188443412335,1,0.8601188443412335}  0.733 \\
32         &\cellcolor[rgb]{0.7625930406799268,1,0.7625930406799268} 20.760 &\cellcolor[rgb]{0.7759374359674885,1,0.7759374359674885}  0.734 &\cellcolor[rgb]{0.99999817389564,1,0.99999817389564}  0.469 &\cellcolor[rgb]{0.917311406449342,1,0.917311406449342} 20.512 &\cellcolor[rgb]{0.9557407280923435,1,0.9557407280923435}  0.731 &\cellcolor[rgb]{0.9999970749339098,1,0.9999970749339098}  0.468 &\cellcolor[rgb]{0.7508435344880633,1,0.7508435344880633} 20.773 &\cellcolor[rgb]{0.7759374359674885,1,0.7759374359674885}  0.734 &\cellcolor[rgb]{0.911108513891689,1,0.911108513891689} 20.527 &\cellcolor[rgb]{0.9180042346834233,1,0.9180042346834233}  0.732 \\
16         &\cellcolor[rgb]{0.9998181813854409,1,0.9998181813854409} 19.868 &\cellcolor[rgb]{0.9994535892357079,1,0.9994535892357079}  0.727 &\cellcolor[rgb]{1.0,1,1.0}  0.477 &\cellcolor[rgb]{0.9999999908309705,1,0.9999999908309705} 19.705 &\cellcolor[rgb]{1.0,1,1.0}  0.725 &\cellcolor[rgb]{0.999999963888073,1,0.999999963888073}  0.474 &\cellcolor[rgb]{0.999868981166906,1,0.999868981166906} 19.854 &\cellcolor[rgb]{0.9972337955057715,1,0.9972337955057715}  0.728 &\cellcolor[rgb]{1.0,1,1.0} 19.690 &\cellcolor[rgb]{0.9999658493272318,1,0.9999658493272318}  0.726 \\

%% file: 5-conclusions_future.tex
In this work, we showed that NeRFs can be naturally extended to hyperspectral imagery.  We collected a dataset, described the special considerations needed to handle hyperspectral data, and presented and evaluated a novel algorithm for creating HS-NeRFs that generalizes to arbitrary wavelength inputs.

We also demonstrated sample applications of hyperspectral NeRFs, including hyperspectral super-resolution and imaging sensor simulation.  Future works include improved performance with architectural and process improvements and application of HS-NeRFs to non-destructively estimate material compositions.

% Future works include reducing the number of images or number of wavelengths required to speed up the dataset capture time, compensating for out-of-focus effects, and better handling of the low resolution of hyperspectral cameras as compared to typical RGB cameras.
% improving the training efficiency of approach 3, and using alternate hyperspectral sensors (\eg line-scanning push-broom where each training ray comes from just 1 line).